\documentclass[letterpaper]{article} 
\usepackage{aaai2026}  
\usepackage{times}  
\usepackage{helvet}  
\usepackage{courier}  
\usepackage[hyphens]{url}  
\usepackage{graphicx} 
\usepackage{natbib}  
\usepackage{caption} 
\usepackage{graphicx} 
\usepackage{subfig}
\usepackage{subfloat}

\usepackage{amsmath}
\usepackage{amssymb}
\usepackage{mathtools}
\usepackage{amsthm}
\newtheorem{theorem}{Theorem} 
\newtheorem{lemma}[theorem]{Lemma}
\newtheorem{definition}[theorem]{Definition}
\usepackage[capitalize,noabbrev]{cleveref}
\usepackage{algorithm}
\usepackage{algorithmic}

\usepackage{newfloat}
\usepackage{listings}
\DeclareCaptionStyle{ruled}{labelfont=normalfont,labelsep=colon,strut=off} 
\floatstyle{ruled}
\newfloat{listing}{tb}{lst}{}
\floatname{listing}{Listing}
\usepackage[switch]{lineno}
\title{Multiple-play Stochastic Bandits with Prioritized Arm Capacity Sharing}
\author{
    Hong Xie\textsuperscript{\rm 1},
    Haoran Gu\textsuperscript{\rm 2},
    Yanying Huang\textsuperscript{\rm 3},
    Tao Tan\textsuperscript{\rm 1},
    Defu Lian\textsuperscript{\rm 1}
}
\affiliations{
    \textsuperscript{\rm 1}University of Science and Technology of China 
    \& State Key Laboratory of Cognitive Intelligence \\
     \textsuperscript{\rm 2}Daqing Oilfield Chongqing Company \\
      \textsuperscript{\rm 3} College of Computer Science, Chongqing University  
}

\usepackage{bibentry}

\begin{document}

\maketitle

\begin{abstract}
This paper proposes a variant of  
multiple-play stochastic bandits tailored to 
resource allocation problems arising from LLM applications, 
edge intelligence applications, etc.  
The proposed model is composed of $M$ arms and $K$ plays.  
Each arm has a stochastic number of capacities, 
and each unit of capacity is associated with a reward function.  
Each play is associated with a priority weight.  
When multiple plays compete for the arm capacity, 
the arm capacity is allocated in a larger priority weight first  manner.  
Instance independent and instance dependent regret lower bounds of 
$\Omega( \alpha_1 \sigma \sqrt{KM T} )$ and 
$\Omega(\alpha_1 \sigma^2 \frac{M}{\Delta} \ln T)$  
are proved,  
where $\alpha_1$ is the largest priority weight and $\sigma$ 
characterizes the reward tail.  
When model parameters are given, 
we design an algorithm named \texttt{MSB-PRS-OffOpt} 
to locate the optimal play allocation policy with a computational complexity   
of $O(M^3K^3)$.   
Utilizing \texttt{MSB-PRS-OffOpt} as a subroutine, 
an approximate upper confidence bound (UCB) based algorithm is 
designed, which has instance independent and instance dependent 
regret upper bounds 
matching the corresponding 
lower bound up to factors of $ \sqrt{K \ln KT }$ and $\alpha_1 K^2$ respectively.    
To this end, we address nontrivial technical challenges arising from 
optimizing and learning under a special nonlinear combinatorial utility function 
induced by the prioritized resource sharing mechanism.   
\end{abstract}


 \section{Introduction}
\label{sec:intro}

The Multi-play multi-armed bandit (MP-MAB) 
is a classical sequential learning framework \cite{anantharam1987asymptotically}. 
The canonical MP-MAB model consists of  
one decision maker who pulls multiple arms per decision round.  
Each pulled arm generates a reward, 
which is drawn from an unknown probability distribution.  
The objective is to maximize the cumulative reward facing the 
exploration vs. exploitation dilemma.   
MP-MAB frameworks are applied to various applications such as 
online advertising \cite{lagree2016multiple,komiyama2017position,Yuan2023}, 
power system \cite{Lesage2017}, 
mobile edge computing \cite{Chen2022,Wang2022,Xu2023}, etc.   
Recently, various variants of MP-MAB were studied, 
which tap potentials of MP-MAB framework for resource allocation problems 
and advance the bandit learning literature 
\cite{Chen2022,Moulos2020,Xu2023,Wang2022,Yuan2023}.  

This paper extends MP-MAB to capture the 
prioritized resourcing sharing mechanism, 
contributing a fine-grained resource allocation model.   
We aim to reveal fundamental insights on the interplay of this mechanism and learning.   
Prioritized resourcing sharing mechanisms are implemented in  
a large class of resource allocation problems arising from 
mobile edge computing \cite{chen2021distributed,gao2022combination,Ouyang2019,Ouyang2023}, ride sharing \cite{Chen2022}, etc., 
and have the potential to enable differentiated services in LLM applications.  
For example, in LLM applications, reasoning tasks and 
LLM instances can be modeled as plays and arms respectively.   
Multiple LLM reasoning tasks (plays) share an instance of LLM (an arm) according to 
their priority quantified by price, membership hierarchy, etc.  
In mobile edge computing systems, the infrastructure of edge intelligence, 
tasks and edge servers can be modeled as plays 
and arms respectively.  
When multiple tasks (or plays) are offloaded to the same edge server (or arm), 
the available computing resource is shared among them 
according to the differentiated pricing mechanism 
(an instance of prioritized resourcing sharing mechanism).  

Formally, we proposed 
MSB-PRS 
(\underline{M}ultiple-play \underline{S}tochastic 
\underline{B}andits with \underline{P}rioritized
\underline{R}esource \underline{S}haring).    
The MSB-PRS is composed of $K \in \mathbb{N}_+$ plays and 
$M\in \mathbb{N}_+$ arms.  
Each play has a priority weight and movement costs.  
Each arm has a stochastic number of units of capacities.  
Plays share the capacity in a high priority weight first manner.  
A play receives a reward scaled by its weight 
only when it occupies one unit of capacity.    
The objective is to maximize the cumulative utility (rewards minus costs) 
in $T \in \mathbb{N}_+$ rounds.  
Some recent works tailored MP-MAB to the same or similar applications  
\cite{Chen2022,Xu2023,Wang2022,Wang2022a,Yuan2023}.  
The key difference to this line of research is 
on the stochastic capacity with bandit feedback  
and prioritized capacity sharing.  
This difference poses new challenges.  
One challenge lies in locating the optimal play allocation policy.  
The movement cost and the prioritized capacity sharing 
impose a nonlinear combinatorial structure on the utility function,  
which hinders locating the optimal allocation.  
In contrast to previous works \cite{Xu2023,Wang2022,Wang2022a}, 
top arms do not warrant optimal allocation.  
This nonlinear combinatorial structure also 
makes it difficult to distinguish optimal allocation from 
sub-optimal allocation from feedback.  
As a result, it is nontrivial to balance the exploring vs. exploitation tradeoff.
We address these challenges.     

\subsection{Contributions} 

{\bf Model and fundamental learning limits.}  
We formulate MSB-PRS, which captures the 
prioritized resourcing sharing nature of resource allocation problems.  
We prove instance independent and instance dependent regret lower bounds of 
$\Omega( \alpha_1 \sigma \sqrt{KM T} )$ and 
$\Omega(\alpha_1 \sigma^2 \frac{M}{\Delta} \ln T)$ respectively.  
Technically, we tackle the aforementioned 
nonlinear combinatorial structure challenge 
by identifying one special instances  
of the MSB-PRS that are composed of carefully designed multiple 
independent groups of classical multi-armed bandits and batched MP-MAB.  

{\bf Efficient learning algorithms.} 
{\bf (1) Computational efficiency.} 
Given model parameters, 
to tackle the computational challenge of locating the optimal play allocation 
policy, 
we characterize the aforementioned nonlinear combinatorial structure 
by constructing a priority ranking aware bipartite graph.  
A connection between the utility of arm allocation policies and 
the saturated, monotone and priority compatible matchings is established.    
This connection enables us to design \texttt{MSB-PRS-OffOpt}, 
which locates the optimal play allocation policy with a complexity $O(M^3K^3)$ from a search space with size $K^M$.    
{\bf (2) Sample efficiency.}  
Utilizing \texttt{MSB-PRS-OffOpt} as a subroutine, 
we design an approximate UCB based algorithm, 
which reduces the per-round computational complexity of the exact UCB based algorithm from $K^M$ to $O(K^3M^3)$.  
We prove sublinear instance independent and instance dependent regret upper bounds 
matching the corresponding 
lower bounds up to factors of $ \sqrt{K \ln KT }$ and $\alpha_1 K^2$ respectively.  
The key proof idea is exploiting the monotone property of the utility function to: 
(1) prove the validity of the approximate UCB index; 
(2) show suboptimal allocations make progress in 
improving the estimation accuracy of poorly estimated parameters,  
which gear the learning algorithm toward identifying 
more favorable play allocation policies.

\section{Related Work}
\label{sec:relatedWork}
 
Anantharam \textit{et al.} \cite{anantharam1987asymptotically} 
proposed the canonical MP-MAB model, where  
they established an asymptotic lower bound on the regret and designed
an algorithm achieving the lower bound asymptotically.  
Komiyama \textit{et al.} \cite{komiyama_optimal_2015} 
showed that Thompson sampling achieves the 
regret lower bound in the finite time sense.  
Anantharam \textit{et al.} \cite{anantharam1987b} extended the 
canonical MP-MAB model from 
IID rewards to Markovian rewards.  
This Markovian MP-MAB model was further extended to the rested bandit setting 
\cite{Moulos2020}. 
MP-MAB with a reward function depending on the order of plays 
was studied in \cite{lagree2016multiple,komiyama2017position}. 
This reward function was motivated 
by clicking the model of web applications.  
They established lower bounds on the regret and designed a 
UCB based algorithm to balance the exploration vs. exploitation tradeoff.  
MP-MAB with switching cost is studied in \cite{Agrawal1990,Jun2004}.  
They proved the lower bound on the regret and designed algorithms 
that achieve the lower bound asymptotically.   
MP-MAB with budget constraint is considered in \cite{Luedtke2019,Xia2016,Zhou2018} 
and a stochastic number of plays in each round is 
considered in \cite{Lesage2017}, which is motivated by power system.   
Recently, Yuan \textit{et al.} \cite{Yuan2023} extended the 
canonical MP-MAB classical to the sleeping bandit setting, 
for the purpose of being tailored to the recommender systems.  

Our work is closely related to \cite{Chen2022,Wang2022,Xu2023}.  
Chen \textit{et al.} \cite{Chen2022} 
tailored the canonical MP-MAB model for the user-centric 
selection problems.  
Their model considered homogeneous plays and 
expert feedback on capacity.  
They designed a Quasi-UCB algorithm for this problem 
with sublinear regret upper bounds.  
Our work generalizes their model to capture heterogeneous plays, 
prioritized resourcing sharing, and bandit feedback on the capacity.  
This extension not only be more friendly to real-world applications, 
but also incurs new challenges for locating the optimal allocation and 
design learning algorithms.  
We design a UCB based algorithm and prove 
both regret upper bounds and lower bounds.  
Wang \textit{et al.} \cite{Wang2022} proposed a model that also allowed 
multiple plays to share capacity on an arm.  
Their model considers a deterministic capacity provision.   
The capacity is unobservable and coupled with the reward.  
They proved regret lower bound on regret and 
designed an action elimination based algorithm whose regret matches 
the regret lower bound to a certain level.  
Xu \textit{et al.} \cite{Xu2023} extended this model to the setting with 
strategic agents and competing for the capacity.  
They analyzed the Nash equilibrium in the offline setting 
and proposed a Selfish MP-MAB with an Averaging Allocation 
approach based on the equilibrium. 

Various works share some connections to 
the MP-MAB research line.  
Combinatorial bandits 
\cite{cesa2012combinatorial,chen2013combinatorial,Combes2015} 
generalize the reward function of 
the canonical MP-MAB from linear to non-linear.  
Various variants of combinatorial bandits were studied: 
(1) combinatorial bandits with semi-bandit feedback \cite{chen2013combinatorial,chen2016combinatorial,gai2012combinatorial,Combes2015},  
i.e., the reward of each pulled arm is revealed; 
(2) combinatorial bandits with bandit feedback: \cite{cesa2012combinatorial,Combes2015}, 
i.e., only one reward associated with the pulled arm set is revealed; 
(3) combinatorial bandits with different combinatorial structures, 
i.e., matroid \cite{kveton2014matroid}, 
$m$-set \cite{anantharam1987asymptotically}, 
permutation \cite{gai2012combinatorial}, etc.    
Cascading bandit 
\cite{combes2015learning,kveton2015combinatorial,wen2017online} 
extends the reward function of 
the canonical MP-MAB from linear to 
a factorization form over the set of selected arms.  
Decentralized MP-MAB (a.k.a. multi-player MAB)~\cite{Agarwal2022multi,anandkumar2011distributed,rosenski2016multi,bistritz2018distributed,wang2020optimal})   considers the setting that players either cannot communicate with others 
or their communication is restrictive.  

 \section{MSB-PRS Model}

\subsection{Model Setting}

For any integer $N$, the notation $[N]$ denotes a set  
$[N] \triangleq \{1,\ldots, N\}$.  
The MSB-PRS consists of one decision maker, 
$M \in \mathbb{N}_+$ arms, $K \in \mathbb{N}_+$ plays 
and a finite number  of $T \in \mathbb{N}_+ $ decision rounds.  
In each decision round $t \in [T]$,  
the decision maker needs to assign all $K$ plays to arms.  
Each play can be assigned to one arm, 
and multiple plays can be allocated to the same arm.  
The objective is to maximize the total utility, 
whose formal definition is deferred after 
the arm model and reward model are made clear.   

{\bf Arm model.}  
The arm $m \in [M]$ is characterized by a pair of random variables 
$(D_m, R_m)$, where $D_m$ characterizes the 
stochastic availability of capacity and 
$R_m$ characterizes the per unit capacity rewards.  
The support of $D_m$ is a subset of $[d_{\max}]$, 
where $d_{\max} \in \mathbb{N}_+$ denotes the 
maximum possible units of capacity on an arm.  
Let $D^{(t)}_{m}$ denote the number of units of capacity available 
on arm $m$ in round $t$.  The $D^{(t)}_{m}$ is drawn from $D_m$, 
i.e., 
$
D^{(t)}_{m} \sim D_m,
$
and each $D^{(t)}_{m}$ drawn from $D_m$ is independent across $t$ and $m$.  
The $i$-th unit of  capacity  on arm $m$ is associated with a reward denoted by 
$R^{(t)}_{m, i}$, where $i \in [D^{(t)}_{m}]$.  
The $R^{(t)}_{m, i}$ is drawn from $R_m$, 
i.e., 
$
R^{(t)}_{m, i} \sim R_m
$
whose support is a subset of $\mathbb{R}$, 
and each $R^{(t)}_{m, i}$ drawn from $R_m$ is independent across $t, m$ and $i$.
Denote the mean of $R_m$ as
$
\mu_m \triangleq \mathbb{E} [R_m].  
$
Without loss of generality, we assume $\mu_m >0, \forall m \in [M]$.  
We assume that  $R_m$  is $\sigma$-subgaussian, 
where $\sigma \in \mathbb{R}_+$.  
Let $\boldsymbol{\mu}\triangleq [\mu_{m}:\forall m \in [M]] $ denote the reward mean vector. 
Let $\boldsymbol{P}_{m} \triangleq [P_{m,d}:\forall d \in [d_{max}]]$ denote the 
complementary cumulative probability vector of $D_{m}$, where 
	\begin{equation}
	\begin{aligned}
	P_{m,d}&=\mathbb{P}[D^{(t)}_{m} \geq d], \forall d \in [d_{max}],m\in[M].
	\nonumber
        \end{aligned}
	\end{equation}
For presentation convenience, denote the complementary cumulative 
probability matrix as:
	\begin{equation}
	\begin{aligned}
	\boldsymbol{P}\triangleq[P_{m,d}:\forall d \in [d_{max}],m\in[M]].
        \nonumber	
        \end{aligned}
	\end{equation}
The $\boldsymbol{\mu}$ and $\boldsymbol{P}$ are unknown to the decision maker.   
Arms can model instances of LLM, edge servers, etc (refer to Section \ref{sec:intro}).

{\bf Play and priority model.}   
The play $k \in [K]$ is characterized by $(\boldsymbol{c}_k,\alpha_k)$, 
where $\boldsymbol{c}_k \in (\mathbb{R}_+ \cup \{+\infty\})^M$ and $\alpha_k \in \mathbb{R}_+$.  
The $\boldsymbol{c}_{k}$ denotes the movement cost 
vector associated with play $k$ and denote its entries as 
$\boldsymbol{c}_{k} \triangleq [c_{k,m}:\forall m\in[M]]$, 
where $ c_{k,m} \in \mathbb{R}_+\cup \{+\infty\}$ denotes the movement cost of 
assigning play $k$ to arm $m$.   
The case $ c_{k,m} = +\infty$ models the constraint that arm $m$ is unavailable to play $k$.   
The weight $\alpha_k$ quantifies the priority of play $k$.   
Larger weight implies higher priority.  
Without loss of generality, we assume 
\[
\alpha_1 \geq \alpha_2 \geq \cdots \geq \alpha_K > 0. 
\]
The $\alpha_k$'s capture differentiated service of 
mobile edge computing, or the superiority of cars in ride sharing systems. 
Both $\boldsymbol{c}_k$ and $\alpha_k$ are known to the decision maker.  
Plays can model reasoning tasks, computing tasks, etc (refer to Section \ref{sec:intro}).

{\bf Prioritized capacity sharing model.}   
Let $a^{(t)}_{k} \in [M]$ denote the arm pulled by play $ k \in [K] $ in round $t$.   
Denote the play allocation or action profile in round $t$ as
$
\boldsymbol{a}^{(t)}\triangleq[a^{(t)}_{k}:\forall k\in[K]].  
$
Denote the number of plays assigned to arm $m$ in round $t$: 
\[ 
N^{(t)}_{m} \triangleq \sum_{k\in[K]}\boldsymbol{1}_{\{a^{(t)}_{k}=m\}}
\]    
Plays are prioritized according to their weights.  
Specifically, in round $t$, the $N^{(t)}_{m}$ plays assigned to arm $m$ are ranked 
according to their weights, i.e., $\alpha_k$'s, in descending order, 
where ties are broke arbitrarily, and they share the capacity according to this order.    
Consider a play assigned to arm $m$, 
i.e., $a^{(t)}_k = m$, 
denote its rank on arm $m$ as $\ell^{(t)}_{k} \in [K]$.     
In round $t$, only top-$\min\{ N^{(t)}_{m}, D^{(t)}_{m}\}$ plays assigned 
to arm $m$ are allocated capacities,  
in a fashion that one unit of capacity per play.  
Namely, 
when the capacity is abundant, i.e., $D^{(t)}_{m} {\geq} N^{(t)}_{m}$, 
the $D^{(t)}_{m} {-} N^{(t)}_{m}$ units of capacity are left unassigned; 
and when the capacity is scarce, i.e., 
$D^{(t)}_{m} {<} N^{(t)}_{m}$, 
the $N^{(t)}_{m} {-} D^{(t)}_{m}$ plays do not get capacity.


{\bf Rewards and feedback.} 
Once play $k$ gets a unit of capacity, a reward scaled by the weight is generated: 
\[
X^{(t)}_{k} \triangleq
\left\{ 
\begin{aligned}
& 
\alpha_k R^{(t)}_{m,\ell^{(t)}_{k}}, 
&& \text{if } \ell^{(t)}_{k} \leq D^{(t)}_{a^{(t)}_k},
\\
& 
\text{null}, 
&& \text{otherwise}.  
\end{aligned}
\right.
\]
where null models that play $k$ does not receive any reward 
when it does not occupy any capacity.  
The decision maker observes the rewards received by each arm.  
Let 
\[
\boldsymbol{X}^{(t)} \triangleq[X^{(t)}_{k}:\forall k\in [K]]
\]
denote the reward vector 
observed in round $t$.  
In round $t$, the number of capacity $D^{(t)}_{m}$ is revealed to the decision maker 
if and only if at least one play is assigned to arm $m$ in this round, i.e., $N^{(t)}_{m}>0$. 
Denote the capacity feedback vector 
\[
\boldsymbol{D}^{(t)} 
{\triangleq} [ D^{(t)}_{m} : m \in \{ m' | N^{(t)}_{m'} {>} 0 \} ].
\]
The decision maker observes $\boldsymbol{X}^{(t)}$ and $\boldsymbol{D}^{(t)}$ in round $t$.  

\subsection{Problem Formulation}

Denote the expected total reward generated from arm $m$ 
in round $t$ as 
$\overline{R}_{m}(\boldsymbol{a}^{(t)}; \mu_m, \boldsymbol{P}_m)$, formally: 
	\begin{align*}
	\overline{R}_{m}(\boldsymbol{a}^{(t)}; \mu_m, \boldsymbol{P}_m) 
	& 
	\triangleq 
      \mathbb{E} 
      \left[ 
      \sum_{k \in [K]} \boldsymbol{1}_{\{ a^{(t)}_{k} = m \}} X^{(t)}_{k}
      \right]
\\
&
= \mu_m 
      \sum_{k \in [K]} 
      \boldsymbol{1}_{\{ a^{(t)}_{k} = m \}} \alpha_k P_{m,\ell^{(t)}_{k}}. 
        \nonumber
 	\end{align*} 
Let $U_{m}(\boldsymbol{a}^{(t)}; \mu_m, \boldsymbol{P}_m)$ denote the expected utility 
earned from arm $m$ in round $t$.  
It is defined as the expected reward minus  the movement cost, 
formally:  
\begin{align*}
& 
U_{m}(\boldsymbol{a}^{(t)}; \mu_m, \boldsymbol{P}_m) 
\\
&	
\triangleq \overline{R}_{m}(\boldsymbol{a}^{(t)} ; \mu_m, \boldsymbol{P}_m) 
	- \sum_{k\in [K]} c_{k,m} \boldsymbol{1}_{ \{a^{(t)}_{k}=m\}}.
        \nonumber
\end{align*}
Let $U(\boldsymbol{a}^{(t)}; \boldsymbol{\mu}, \boldsymbol{P})$ denote the aggregate utility from all plays given action profile 
$ \boldsymbol{a}_{t} $, formally: 
	\begin{equation}
	\begin{aligned}
	U(\boldsymbol{a}^{(t)}; \boldsymbol{\mu}, \boldsymbol{P})  
	\triangleq \sum_{m\in[M]} U_{m}(\boldsymbol{a}^{(t)}; \mu_m, \boldsymbol{P}_m)  .
	\end{aligned}
    \label{eq_U(a)}
	\end{equation}
The objective is to maximize the total utility in $T$ rounds, 
i.e., maximize $\sum_{t=1}^{T} U(\boldsymbol{a}^{(t)}; \boldsymbol{\mu}, \boldsymbol{P})$. 
Since the system is stationary in $t$, 
the optimal action profile across different time slots 
can be expressed as: 
	\begin{equation}\label{eq:model:OptAction}
	\begin{aligned}
	\boldsymbol{a}^{*}  
	\in \arg\max_{\boldsymbol{a}\in \mathcal{A}} U(\boldsymbol{a}; \boldsymbol{\mu}, \boldsymbol{P}).
	\end{aligned}
	\end{equation}
where $\mathcal{A} \triangleq [M]^{K}$. 
Note that $\boldsymbol{a}^{*} $ is unknown to the decision maker 
because the parameters $\boldsymbol{\mu}$ and $\boldsymbol{P}$ are unknown. 
We define the regret as: 
\[
\text{Reg}_T 
\triangleq 
\mathbb{E} 
\left[
\sum_{t=1}^T 
\left( 
U(\boldsymbol{a}^\ast; \boldsymbol{\mu}, \boldsymbol{P}) 
- 
U(\boldsymbol{a}^{(t)}; \boldsymbol{\mu}, \boldsymbol{P})
\right)
\right].  
\]

{\bf Remark:} 
The utility function $U(\boldsymbol{a}^{(t)}; \boldsymbol{\mu}, \boldsymbol{P}) $ has a 
nonlinear combinatorial structure with respect to 
$\boldsymbol{\mu}, \boldsymbol{P}$ and it has a cost term.  
As a consequence, arms with large per unit rewards are not necessarily favorable.  
There are in total $|\mathcal{A}|= M^{K}$ action profiles.  
Thus, locating $\boldsymbol{a}^{*}$ is nontrivial.    
Distinguishing optimal action profile from 
sub-optimal allocation from feedback is not easy.  
It is nontrivial to tackle this nonlinear combinatorial structure 
to reveal fundamental 
learning limits and balance the exploring vs. exploitation tradeoff.

\section{Fundamental Learning Limits}

We reveal fundamental limits of learning the optimal action profile by 
proving instance independent 
and instance dependent regret lower bounds.  
\begin{theorem}  
For any learning algorithm, 
there exists an instance of MSB-PRS such that  
\[
\text{Reg}_T \geq 
\frac{1}{27} \alpha_1 \sigma \sqrt{ MKT}.    
\]   
Furthermore, 
$\text{Reg}_T \geq \Omega(\alpha_1 \sigma \sqrt{MK T})$. 
\label{thm:InsIndLower}
\end{theorem}

The key proof idea  
is identifying one special instances  
of the MSB-PRS that are composed of carefully designed multiple 
independent groups classical multi-armed bandits and batched MP-MAB. 
For each group, we apply Theorem 15.2 of \cite{lattimore2020bandit} 
to bound its regret lower bound.  
Finally, summing them up across groups 
we obtain the instance independent regret lower bound. 
 
\begin{theorem}
Consider $\Delta \in \mathbb{R}_+$ utility gap MSB-PRS, 
i.e., the class of MSB-PRS satisfy 
\[ 
U(\boldsymbol{a}^\ast; \boldsymbol{\mu}, \boldsymbol{P}) 
-
\max_{
\boldsymbol{a}: 
U(\boldsymbol{a}; \boldsymbol{\mu}, \boldsymbol{P}) 
\neq 
U(\boldsymbol{a}^\ast; \boldsymbol{\mu}, \boldsymbol{P})
} 
U(\boldsymbol{a}; \boldsymbol{\mu}, \boldsymbol{P}) 
= 
\Delta.  
\]
For any consistent play allocation algorithm, 
there exists $\Delta$ utility gap instances of MSB-PRS, such that 
the regret of any consistent learning algorithm on them satisfies
\[
\liminf_{T\rightarrow \infty} 
\frac{\text{Reg}_T}{\ln T} 
\geq 
 \alpha_1 \sigma^2  \frac{ M}{\Delta}.
\] 
\label{thm:InsDenLower}
\end{theorem}
The idea of restricting to $\Delta$ utility gap MSB-PRS 
in proving the instance dependent regret lower bound 
follows the work \cite{Kveton2015}, 
which proves the instance dependent regret lower bound of 
stochastic combinatorial semi-bandits restricting to 
gap instances, instead of the basic model parameters.  
The proof routine is similar to that of the Theorem \ref{thm:InsIndLower}.  
The constructed special instances of the MSB-PRS 
are nearly the same as that of Theorem \ref{thm:InsIndLower}, 
except that for each group of bandits, 
we carefully design their mean gap, 
such that the total gap equals $\Delta$.  
We apply Theorem 16.2 of \cite{lattimore2020bandit} to bound 
the asymptotic lower bound.  
Finally, summing them up across groups 
we obtain the instance dependent regret lower bound.

\section{\bf Efficient Learning Algorithms}
\label{sec:offline}

\subsection{\bf Efficient Computation Oracle}
\label{sec:BiparGraph}

Given model parameters, 
we design \texttt{MSB-PRS-OffOpt} to locate the optimal action profile, 
which will serve as an efficient computation oracle 
for learning the optimal action profile. 

{\bf Bipartite graph formulation.}  
We formulate a complete weighted bipartite graph with node set 
$\mathcal{U} \cup \mathcal{V}$ 
and edge set $\mathcal{U} \times \mathcal{V}$, 
where $\mathcal{U} \cap \mathcal{V} = \emptyset$ and 
\begin{align*}
& 
\mathcal{U} \triangleq \{u_1, \ldots, u_K\},  
\quad
\mathcal{V} 
\triangleq 
\bigcup_{ m \in [M] } 
\mathcal{V}_m,  
\\
&
\mathcal{V}_m 
\triangleq 
\{
v_{m,1}, \ldots, v_{m,K} 
\}.  
\end{align*} 
The node $u_k \in \mathcal{U}$ 
corresponds to play $k \in [K]$.  
The node set $\mathcal{V}_m$  
corresponds to arm $m \in [M]$.     
Nodes $v_{m,j} \in \mathcal{V}_m$, 
where $j \in [K]$,   
are designed to capture the prioritized resource sharing mechanism.   

Denote $\Lambda_m (k, \ell)$ as the marginal utility contribution of play 
$k$ on an arm when it is ranked $\ell$-th among all plays pulling this arm, 
formally  
\[
\Lambda_m (k, \ell)
\triangleq   
\alpha_k  \mu_m P_{m, \ell} - c_{k,m}.  
\]
The $U_{m}(\boldsymbol{a}; \mu_m, \boldsymbol{P}_m) $ can be decomposed as: 
\[
U_{m}(\boldsymbol{a}; \mu_m, \boldsymbol{P}_m) 
=
\sum_{k \in [K]}
\boldsymbol{1}_{\{a_k = m\}}  
\Lambda_m (k, \ell_k (\boldsymbol{a})), 
\]
where $\ell_k (\boldsymbol{a})$ denote the rank of play 
on the arm $a_{k}$ according to the prioritized capacity 
sharing mechanism.  
Denote $\theta_k$ as the number of plays proceeding 
play $k$ with respect to their priority weights 
\[
\theta_k \triangleq 
|
\{
k' | 
\alpha_{k'} \geq \alpha_k
\}
|.  
\]
The prioritized capacity sharing mechanism implies 
an upper bound on the rank of $k$, i.e.,   
$
\ell_k( \boldsymbol{a} ) 
\leq \theta_k.  
$
Namely, on each arm, play $k$ would be ranked at most 
$\theta_k$-th regardless of the number of plays assigned to this arm.  
 
Denote a weight function over the edge set as: 
$
W: \mathcal{U} \times \mathcal{V} \rightarrow \mathbb{R}.     
$    
The weight of the edge $(u_k, v_{m,j})$ is defined as: 
\[
W( u_k, v_{m,j} ) 
= 
\left\{
\begin{aligned}
& 
\Lambda_m({k}, j), 
&& 
\text{ if $j \leq \theta_k$}, 
\\
& 
- \infty, 
&& 
\text{ otherwise}.
\end{aligned}
\right.
\]
The weight $W( u_k, v_{m,j} )$ quantifies the marginal utility 
contribution of play $k$ for pulling arm $m$, 
when it is ranked $j$-th.    
As imposed by the prioritized capacity sharing mechanism, 
the rank of play $k$ can not exceed $\theta_k$.  
We thus set the utility associated with such invalid rank 
as $- \infty$ to disable these edges.  
Denote the weighted bipartite graph as 
\[
G = (\mathcal{U} \cup \mathcal{V}, \mathcal{U} \times \mathcal{V}, W).  
\]

{\bf From action profiles to matchings}.  
Let $\mathcal{M} \subseteq \mathcal{U} \times \mathcal{V}$ 
denote a matching in graph $G$, 
which is a set of pairwise non-adjacent edges, i.e.,  
$|\{u | (u,v) \in \mathcal{M} \}| = |\{v | (u,v) \in \mathcal{M} \}| 
= | \mathcal{M} |$.   
Denote the index of the arm that is linked to node $v_{m,j}$ under 
$\mathcal{M}$ as 
\[
\phi_{m,j} (\mathcal{M})
\triangleq  
\left\{
\begin{aligned}
& k, 
&& 
\text{if $(u_k, v_{m,j}) \in \mathcal{M}$},
\\
& 
\text{0}, 
&& 
\text{otherwise}, 
\end{aligned}
\right.
\]
where index $0$ is defined as a dummy play 
and we define its weight is as $\alpha_0 = 0$.  
Denote an indicator function associated with 
$\mathcal{M}$ as: 
\[
b_{m,j} (\mathcal{M})
\triangleq  
\left\{
\begin{aligned}
& 1, 
&& 
\text{if $\exists k, (u_k, v_{m,j}) \in \mathcal{M}$}, 
\\
& 
\text{0}, 
&& 
\text{otherwise.}
\end{aligned}
\right.
\]

We next define a class of matchings that 
can be connected to the action profiles.  

\begin{definition}
A matching $\mathcal{M}$ 
is: 
(1) $\mathcal{U}$-saturated if   
$
\{
u | (u,v) \in \mathcal{M}
\} 
= \mathcal{U} 
$; 
(2) $\mathcal{V}$-monotone if 
$b_{m,j} (\mathcal{M}) \geq b_{m,j'} (\mathcal{M}), 
\forall j < j'$; 
(3) priority compatible  if  
$\alpha_{\phi_{m,j} (\mathcal{M})} \geq \alpha_{\phi_{m,j'} (\mathcal{M})}, 
\forall j < j'
$. 
 \end{definition}

The $\mathcal{U}$-saturated property states that 
each play node is an endpoint of one edge  
of $\mathcal{M}$.      
The $\mathcal{V}$-monotone property states that end points of $\mathcal{M}$ 
on the $\mathcal{V}_m$ side 
forms an increasing set, 
i.e., it can be expressed as 
$
\{
v_{m,1}, \ldots, v_{m,J}
\}
$, 
where $J = | 
\{
v | (u,v) \in \mathcal{M} 
\}
\cap 
\mathcal{V}_m|$. 
 
\begin{lemma}
Action profile $\boldsymbol{a} {\in} \mathcal{A}$ can be mapped into 
a $\mathcal{U}$-saturated, $\mathcal{V}$-monotone, and 
priority compatible matching 
$
\widetilde{\mathcal{M}} (\boldsymbol{a}) 
= 
\left\{
(u_k, v_{a_k, \ell_k (\boldsymbol{a}) }) | 
k \in [K] 
\right\}. 
$  
Furthermore, 
it holds that 
$
U(\boldsymbol{a}; \boldsymbol{\mu}, \boldsymbol{P}) 
= \sum\nolimits_{ (u,v) \in \widetilde{\mathcal{M}}(\boldsymbol{a}) } 
W(u,v),
$
and $\widetilde{\mathcal{M}} (\boldsymbol{a}) \neq \widetilde{\mathcal{M}} (\boldsymbol{a}')$ 
for any $\boldsymbol{a} \neq \boldsymbol{a}'$.   
\label{lem:offline:ActionToMatching}
\end{lemma}  

Lemma \ref{lem:offline:ActionToMatching} states that each action profile 
can be mapped into a $\mathcal{U}$-saturated 
, $\mathcal{V}$-monotone and priority compatible matching 
with utility equals the weights of the matching.

{\bf From matchings to action profiles.}    
In the following lemma, we show that 
a $\mathcal{U}$-saturated, $\mathcal{V}$-monotone, 
and priority compatible matching 
can be mapped into an action profile 
with weights of the matching equals the utility.

\begin{lemma}
A $\mathcal{U}$-saturated, $\mathcal{V}$-monotone, and 
priority compatible matching $\mathcal{M}$ 
can be mapped into action profile  
$
\widetilde{\boldsymbol{a}}(\mathcal{M}) 
\triangleq 
(\widetilde{a}_k (\mathcal{M}) : 
\forall k \in [K])$, 
where   
\[
\widetilde{a}_k (\mathcal{M})
=\sum_{m \in [M]} m
\sum_{j \in [K]} 
\boldsymbol{1}_{ \{ \phi_{m,j} (\mathcal{M}) = k \}}.
\]
Furthermore, 
$
U( \widetilde{\boldsymbol{a}} (\mathcal{M}), \boldsymbol{\mu}, \boldsymbol{P}) 
= \sum\nolimits_{ (u,v) \in \mathcal{M} } 
W(u,v).  
$
\label{lem:offline:MatchingToAction}
\end{lemma}

{\bf Locating the optimal action profile.}  
Lemma \ref{lem:offline:ActionToMatching} and 
\ref{lem:offline:MatchingToAction} imply that 
locating the optimal action profile 
is equivalent to searching the 
$\mathcal{U}$-saturated, $\mathcal{V}$-monotone, and priority compatible 
matching with the maximum total weights.   
However, a maximum weighted matching may not be 
$\mathcal{U}$-saturated, $\mathcal{V}$-monotone and priority compatible.  
This hinders one to apply the maximum weighted matching algorithm.  
For any $\mathcal{U}$-saturated matching $\mathcal{M}$,   
if it is not $\mathcal{V}$-monotone or priority compatible, 
it can be adjusted to be  a
$\mathcal{V}$-monotone and priority compatible matching 
$\mathcal{M}'$: 
\begin{align}
\mathcal{M}' 
\triangleq 
\bigcup_{m \in [M]} 
\bigcup_{j=1}^{|\mathcal{K}_m|} 
\{
(u_{L_{m,j}}, v_{m,j})
\}
\label{eq:OffOpt:adjust}
\end{align}
where $\mathcal{K}_m = \{j | \phi_{m,j} (\mathcal{M}) \neq 0\}$ denotes 
a set of plays linked to arm $m$ by $\mathcal{M}$, 
 $L_{m,1}, \ldots, L_{m, |\mathcal{K}_m|}$ is a ranked list 
of $\mathcal{K}_m$ such that 
$L_{m,j} < L_{m,j'}, \forall j < j'$. 
Furthermore, it can be easily verified that 
$
\sum_{ (u,v) \in \mathcal{M} }  
W(u,v) 
\leq 
\sum_{ (u,v) \in \mathcal{M}' }  
W(u,v)
$.  
The implication is that one can first locate the 
maximum weighted matching 
(the maximum weighted matching is $\mathcal{U}$-saturated). 
If it does not have all three desired properties, 
one can apply the above strategy to adjust it to 
have all three desired properties. 
Locating the maximum weight matching 
is a well studied problem.  
The Hungarian algorithm and its variants such as Crouse \textit{et al. \cite{7738348}}  provide 
computationally efficient algorithms for this problem.  
Algorithm \ref{alg:OptimalWithHungarian} combines the above elements 
to locate the optimal action profile.  
The essential computational complexity is the maximum weighted matching.  
The computational complexity of Algorithm \ref{alg:OptimalWithHungarian} 
is $O(M^3 K^3)$, if 
Crouse \textit{et al. \cite{7738348}}  is applied.   

\begin{algorithm}[h]
\caption{\texttt{MSB-PRS-OffOpt} $( \boldsymbol{\mu}, \boldsymbol{P})$}
\label{alg:OptimalWithHungarian}
\begin{algorithmic}[1] 
\STATE $G \leftarrow (\mathcal{U} \cup \mathcal{V}, \mathcal{U} \times \mathcal{V}, W)$
\STATE $\mathcal{M} \leftarrow \texttt{MaximumWeightedMatching(G)}$ 
\STATE If $\mathcal{M}$ does not have three desired properties, 
adjust it according to Eq. (\ref{eq:OffOpt:adjust})
\STATE 
$\widetilde{a}_k (\mathcal{M})
\gets \sum_{m \in [M]} m
\sum_{j \in [K]} 
\boldsymbol{1}_{ \{ \phi_{m,j} (\mathcal{M}) = k \}}
$
\STATE {\bf Return:} $\widetilde{\boldsymbol{a}}(\mathcal{M}) =[\widetilde{a}_k(\mathcal{M}) : k \in [K]]$
\end{algorithmic}
\end{algorithm} 

\subsection{Efficient Learning Algorithm}

{\bf Approximate UCB based algorithm.}  
Note that in time slot $t+1$, 
the decision maker has access to the historical feedback up to time slot $t$, formally   
$
\mathcal{H}_t 
\triangleq
(\boldsymbol{D}^{(1)},\boldsymbol{X}^{(1)},\boldsymbol{a}^{(1)},
\ldots,
\boldsymbol{D}^{(t)},\boldsymbol{X}^{(t)},\boldsymbol{a}^{(t)}
)
.
$
Denote the complementary cumulative probability matrix 
estimated from $\mathcal{H}_t$ as 
$
\widehat{\boldsymbol{P}}^{(t)} 
\triangleq 
[ 
\widehat{P}^{(t)}_{m,d} : 
m \in [M], 
d \in [d_{max}]
],
$
where the $\widehat{P}^{(t)}_{m,d}$ is the empirical average:  
\begin{align}
\widehat{P}^{(t)}_{m,d} 
\triangleq
\frac
{\sum_{s=1}^t 
\boldsymbol{1}_{\{N^{(s)}_{m} \geq 1\}}
\boldsymbol{1}_{\{D^{(s)}_{m} \geq d\}}
}
{
\sum_{s=1}^t 
\boldsymbol{1}_{\{N^{(s)}_{m} \geq 1\}} 
}.  
\label{eq:OnlineAlgo:EstP}
\end{align}

Denote the mean vector estimated from $\mathcal{H}_t$ as 
$
\widehat{ \boldsymbol{\mu} }^{(t)} 
= 
[
\widehat{\mu}^{(t)}_{m}  : 
m \in [M]
], 
$
where the $\widehat{\mu}^{(t)}_{m}$ is the empirical average: 
\begin{align}
\widehat{\mu}^{(t)}_{m} 
\triangleq 
\frac
{ 
\sum_{s=1}^t \sum_{k=1}^K  
\boldsymbol{1}_{\{ X^{(s)}_{k}\neq\text{null}\}} 
\boldsymbol{1}_{\{a^{(s)}_{k}=m\}}
X^{(s)}_{k} / \alpha_{k} 
}
{
\sum_{s=1}^t \sum_{k=1}^K 
\boldsymbol{1}_{\{ X^{(s)}_{k}\neq\text{null}\}} 
\boldsymbol{1}_{\{a^{(s)}_{k}=m\}}
}. 
\label{eq:OnlineAlgo:EstMu}
\end{align}

The following lemma states a confidence band for 
the above estimators.  

\begin{lemma}
The estimators $\widehat{P}^{(t)}_{m,d}$ and $\widehat{\mu}^{(t)}_{m}$ 
satisfy:  
\begin{align*}
& 
\mathbb{P}
\left[
\exists t, m, 
| 
\mu_m
-
\widehat{\mu}^{(t)}_{m}  
|
\geq
\epsilon^{(t)}_{m}
\right]
\leq
2 M \delta,  
\\
&
\mathbb{P} 
\left[
\exists t, m, d,
| 
\widehat{P}^{(t)}_{m,d} 
- 
P_{m,d}
| 
{\geq} 
\lambda^{(t)}_{m}
\right]
\leq 
2 M d_{\max} \delta, 
\end{align*}
where $\delta \in (0,1)$, 
$\epsilon^{(t)}_{m} $ and $\lambda^{(t)}_{m}$ are derived as 
\begin{align*}
& 
\epsilon^{(t)}_{m} 
=
\left\{ 
\begin{aligned}
& 
\sqrt{
2
\sigma^2
(\widetilde{n}^{(t)}_m+1)
\ln \frac{ \sqrt{\widetilde{n}^{(t)}_m +1} }{ \delta }
}
\frac{1}{\widetilde{n}^{(t)}_m}, 
&& 
\text{if $\widetilde{n}^{(t)}_m \geq 1$}, 
\\
& 
+ \infty, 
&& 
\text{if $\widetilde{n}^{(t)}_m =0$},
\end{aligned}
\right. 
\\
&
\lambda^{(t)}_{m}
=
\left\{ 
\begin{aligned}
& 
\sqrt{
\frac{ n^{(t)}_{m} + 1 }
{2}
\ln \frac{ \sqrt{ n^{(t)}_{m} + 1} } {\delta}
}
\frac{1}{n^{(t)}_{m}} \wedge 1, 
&& 
\text{if $n^{(t)}_{m} \geq 1$}, 
\\
& 
1, 
&& 
\text{if $n^{(t)}_{m} =0$},
\end{aligned}
\right.   
\end{align*}
where the operation $\wedge$ means selecting 
the smaller value between two, 
$
\widetilde{n}^{(t)}_m {=} {\sum_{s=1}^t \sum_{k=1}^K \boldsymbol{1}_{\{ X^{(s)}_{k}\neq\text{null}\}} \boldsymbol{1}_{\{a^{(s)}_{k}=m\}}}
$
and $n^{(t)}_{m} = \sum_{s=1}^t 
\boldsymbol{1}_{\{N^{(t)}_{m} \geq 1\}}$. 
\label{lem:online:mu}
\end{lemma}

\noindent
For simplicity, we denote 
$\boldsymbol{\epsilon}^{(t)} = [ \epsilon^{(t)}_{m} : m \in [M]]$ 
and $\boldsymbol{\lambda}^{(t)} 
= 
[\lambda^{(t)}_{m}: m\in [M] ]$. 
Based on the above lemma, 
the exact UCB index of action 
profile $\boldsymbol{a}$ can be expressed as: 
\[
\text{Exact-UCB}^{(t)} (\boldsymbol{a})  
= \max_{  \boldsymbol{\mu},\boldsymbol{P},
|\widehat{\mu}^{(t)}_m - \mu_m | \leq \epsilon^{(t)}_m, 
\forall m  
\atop
| \widehat{P}_{m,d}^{(t)} - P_{m,d} | \leq \lambda^{(t)}_{m}, 
\forall m, d
}  U(\boldsymbol{a},\boldsymbol{\mu},\boldsymbol{P}).  
\]
The $\text{Exact-UCB}^{(t)} (\boldsymbol{a})$ has a potential computational issue 
in locating the action profile with larger index.  
Specifically, the $\text{Exact-UCB}^{(t)} (\boldsymbol{a})$ may attain 
the max value at different selections of $\boldsymbol{\mu},\boldsymbol{P}$ 
for different action profiles, 
especially when the confidence band fails.  
In this case, to locate the action profile one can only resort to 
exhaustive search, resulting in a computational complexity of 
$O(K^M)$.   
To avoid this problem, we propose to use the  approximate 
UCB index: 
\begin{align}
\text{UCB}^{(t)} (\boldsymbol{a})
{=}   
U(\boldsymbol{a}, \widehat{ \boldsymbol{\mu} }^{(t)} {+} \boldsymbol{\epsilon}^{(t)},
\widehat{\boldsymbol{P}}^{(t)} 
{+} \boldsymbol{\lambda}^{(t)}).   
\label{eq:estimateActPrfClosedForm} 
\end{align}
One advantage of $\text{UCB}^{(t)} (\boldsymbol{a})$ over 
$\text{Exact-UCB}^{(t)} (\boldsymbol{a}) $ is that all action profile 
share the same parameter 
$\widehat{ \boldsymbol{\mu} }^{(t)} {+} \boldsymbol{\epsilon}^{(t)},
\widehat{\boldsymbol{P}}^{(t)} 
{+} \boldsymbol{\lambda}^{(t)}$.  
Algorithm \ref{alg:OptimalWithHungarian} locates 
the action profile attaining the maximum 
$\text{UCB}^{(t)} (\boldsymbol{a})$ with a computational 
complexity of $O(K^3M^3)$.  
As we shown in the proof of instance independent upper bound, 
the monotonicity of utility function with respect to $\boldsymbol{\mu}$ 
and $\boldsymbol{P}$ element-wisely guarantees 
the UCB validity of $\text{UCB}^{(t)} (\boldsymbol{a})$.
The action profile in round $t$ is then selected by: 
\[
\boldsymbol{a}^{(t)} 
\in 
\arg\max_{\boldsymbol{a} \in \mathcal{A}} 
\text{UCB}^{(t-1)} (\boldsymbol{a}).  
\]
Summarizing the above ideas together, 
Algorithm \ref{alg:UCB} outlines an approximate UCB based algorithm.  

\begin{algorithm}[h]
\caption{\texttt{MSB-PRS-ApUCB} ( $\mathcal{H}_t $) }
\label{alg:UCB}
\begin{algorithmic}[1]
\STATE{$\widehat{P}_{m,d}^{(0)} \leftarrow 1, \widehat{\mu}_{m}^{(0)} \leftarrow 0$}
\FOR{$t=1,\ldots,T$}
\STATE Calculate 
$\epsilon_m^{(t-1)}$ and $\lambda_{m}^{(t-1)}$
applying Lemma (\ref{lem:online:mu})
\STATE 
$\boldsymbol{a}^{(t)} \leftarrow 
\texttt{MSB-PRS-OffOpt}(
\widehat{ \boldsymbol{\mu} }^{(t-1)} {+} \boldsymbol{\epsilon}^{(t-1)}, 
$
$
\widehat{\boldsymbol{P}}^{(t-1)} 
{+} \boldsymbol{\lambda}^{(t-1)})$ 
\STATE  
Observe $\boldsymbol{D}^{(t)}$ and $\boldsymbol{X}^{(t)}$ 
\STATE 
Update $\widehat{P}^{(t)}_{m,d} $  via Eq. (\ref{eq:OnlineAlgo:EstP}), 
$\forall m \in \{ m' | N^{(t)}_{m'} {>} 0 \}$
\STATE 
Update $\widehat{\mu}^{(t)}_{m}$ via Eq. (\ref{eq:OnlineAlgo:EstMu}), 
$\forall m \in \{ m' | N^{(t)}_{m'} {>} 0 \}$
\ENDFOR
\end{algorithmic}
\end{algorithm}

{\bf Regret upper bounds.} 
The following two theorems state the instance independent 
and instance dependent regret lower bound individually.  
 
\begin{theorem} 
The instance independent regret upper bound of 
Algorithm \ref{alg:UCB} can be derived as: 
\begin{align*}
& \text{Reg}_T 
\leq 
2M(1+d_{\max}) K \mu_{\max}  
\\
&  
+
36
 \alpha_1 (\mu_{\max}+1) (2\sigma +1)  \sqrt{MKT}  \sqrt{K \ln KT} 
\end{align*}
Furthermore, 
$
\text{Reg}_T \leq 
O(\alpha_1 \sigma \mu_{max}  \sqrt{KMT} 
 \sqrt{ K
\ln KT 
}
).  
$
\label{thm:online:InsIndRegret}
\end{theorem} 

Compared to the instance independent regret lower bound 
derived in Theorem \ref{thm:InsIndLower}, 
the regret upper bound matches the lower bound 
up to a factor of 
$\sqrt{K} \sqrt{ \ln KT }$. 
The key proof idea is 
via exploiting the monotone property of the utility function to 
prove the validity of the approximate UCB index.

\begin{theorem}
The instance dependent regret upper bound of 
Algorithm \ref{alg:UCB} can be derived as: 
\begin{align*}
\text{Reg}_T 
\leq
& 
96 M K^2 \alpha^2_1 
(2 \sigma +1)^2
\frac{1}{ \Delta } 
\ln  KT  
\\
& 
+
2M(1+d_{\max})  K \mu_{\max}
\end{align*}
Furthermore, 
$\text{Reg}_T \leq O(M K^2 \alpha^2_1 
\sigma^2
\frac{1}{ \Delta } 
\ln  KT )$. 
\label{them:InsdeBound}
\end{theorem}

Compared to the instance dependent regret lower bound 
derived in Theorem \ref{thm:InsDenLower}, 
the regret upper bound matches the lower bound 
up to a factor of 
$\alpha_1 K^2$.   
The key proof idea of tackling the aforementioned 
nonlinear combinatorial structure in the proof is 
via exploiting the monotone property of the utility function to 
show suboptimal allocations make progress in 
improving the estimation accuracy of poor estimated parameters,  
which gear the learning algorithm toward identifying 
more favorable suboptimal allocations. 
Furthermore, group suboptimal action profiles with respect to their gap 
to the optimal action profile, with a double trick on 
determining the desired gap for each group.  

{\bf Discussion on tightness.}  We believe that closing the regret gap is an open problem, 
since MSB-PRS is neither a standard MP-MAB model nor a standard 
combinatorial bandit model. 

\section{Synthetic Experiments}

\subsection{\bf Parameter setting}.  
We consider $M=5$ arms and $K=10$ plays. It is essential to note that we will  systematically vary $M$ and K to assess the performance of our proposed algorithm. 
The probability mass function and the reward distribution is same as Chen \textit{et al.}~\cite{Chen2022}. 
The probability mass function is defined as 

\[
p_{m,d}  
= 
\left\{
\begin{aligned}
& 
\alpha d, 
&&
\text{ if $d \leq \lceil m/2 \rceil$}, 
\\
&
\alpha (m+1-d), 
&& 
\text{ if $\lceil m/2 \rceil < d \leq m$}, 
\\
& 
0, 
&& 
\text{otherwise}
\end{aligned}
\right.
\]
where $
\alpha
=
1/(
\sum_{d=1}^{\lceil{m/2}\rceil }
d
+
\sum_{d=\lceil{m/2}\rceil+1}^{m }
m+1-d)
$ is the normalizing factor.  
The probability function exhibits a shape akin to a normal distribution. Essentially, as the index $m$ increases, there is an expected augmentation in the number of units of resource associated with an arm. This trend arises due to the shifting of probability masses towards larger values of $m$ with the increase in the index $d$.
Each arm's rewards are sampled from Gaussian distributions, i.e., $\boldsymbol{R}_m\sim N(\mu_m,\sigma^2)$, where $\mu_m\in[1,2]$ and $\sigma > 0$.  We examine three cases regarding the reward mean:
\begin{itemize}
\item 
\textbf{Inc-Shape}: $\mu_m = 1+m/M$, the reward mean increases with the index of arm $m$.
\item 
\textbf{Dec-Shape}: $\mu_m = 2-m/M$, the reward mean decreases with the index of arm $m$.
\item 
\textbf{U-Shape}: $\mu_m = 1+| M/2-m |/M$, the reward mean initially decreases and then increases with the index of arm $m$.
\end{itemize}
We designate the movement cost as 
$
c_{k,m}
=
\eta|(k \mod M) - m| / \max\{K,M\}
$, where $\eta \in \mathbb{R}_+$ is a hyper-parameter that 
controls the scale of the cost.  Unless explicitly varied, we adopt the following default parameters: 
$T = 10^4$, $\delta = 1/T$, $K=10$ plays, $M=5$ arms, 
$\eta=1$, $\sigma=0.2$ and the U-Shape reward.  Furthermore, the number of play types is 2, with parameter 
$\alpha=[3,1]$.

\begin{figure}[htb]
	\centering
	\subfloat[Regret vs. $M$ \label{fig_narms:selComp}]
	{
		\includegraphics[width=0.48\linewidth]{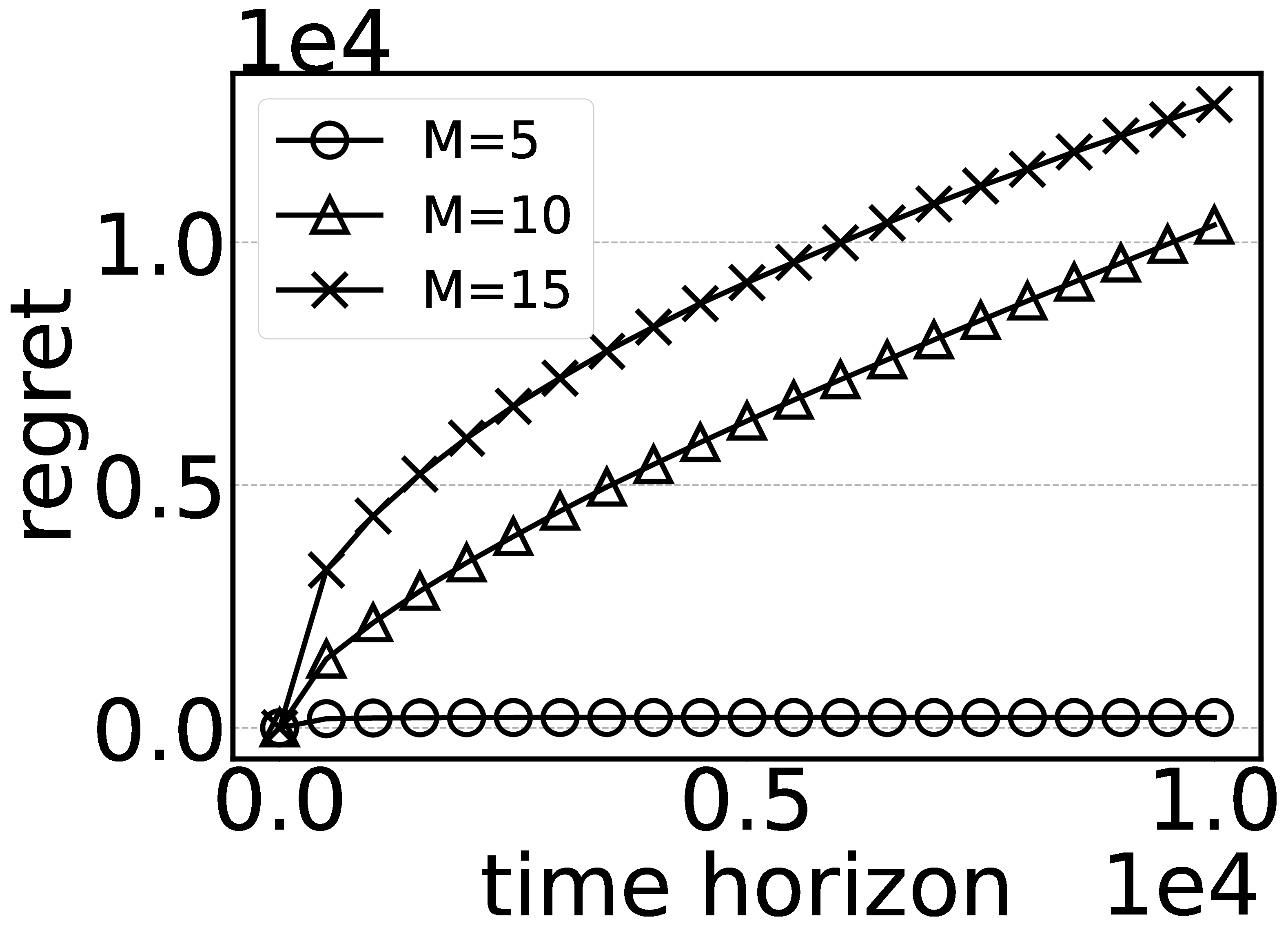}
	}
	\subfloat[ $M=5$ \label{fig_narms:1}]{
		\includegraphics[width=0.46\linewidth]{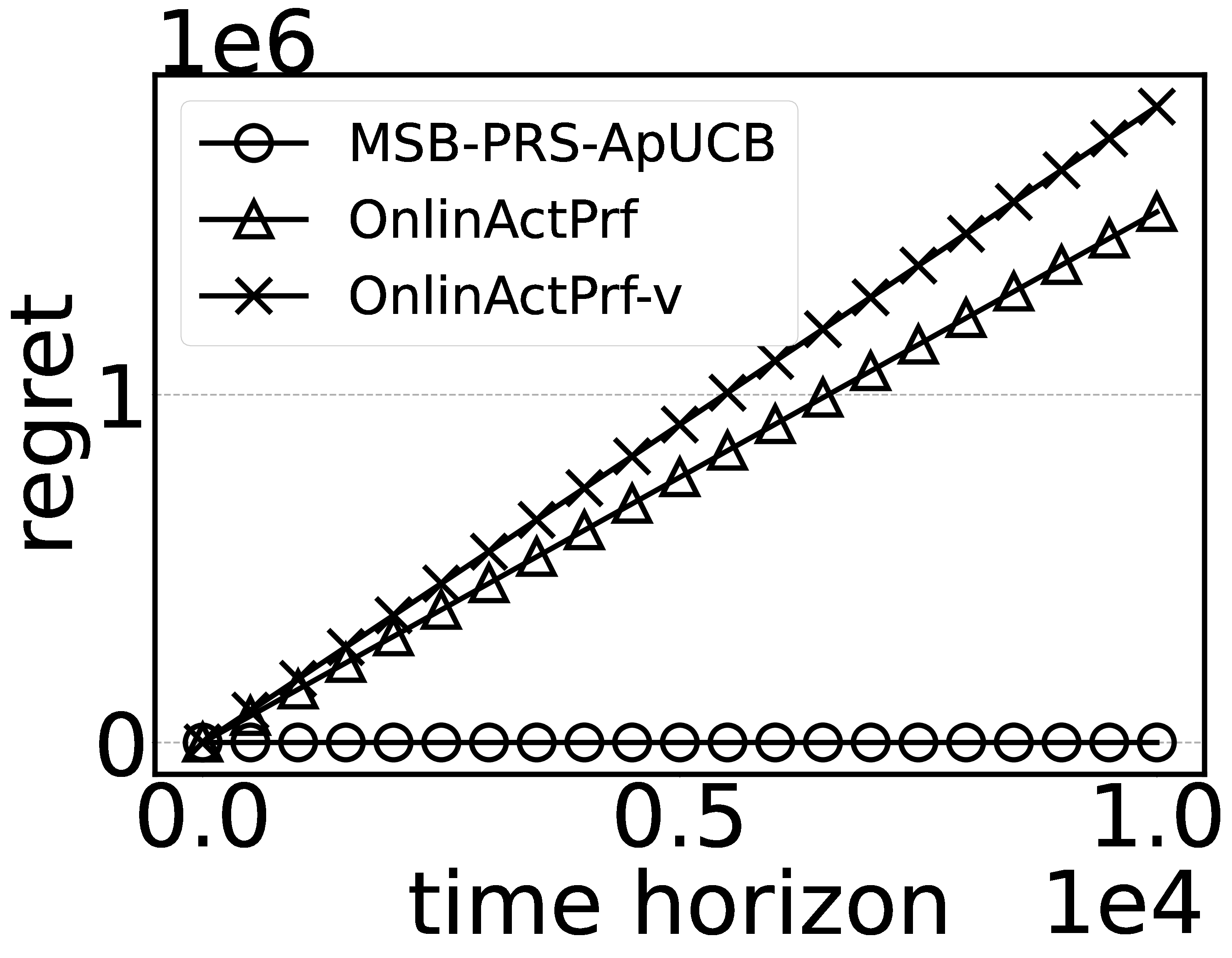}
	}\\
	\subfloat[$M=10$ \label{fig_narms:2}]
	{
		\includegraphics[width=0.48\linewidth]{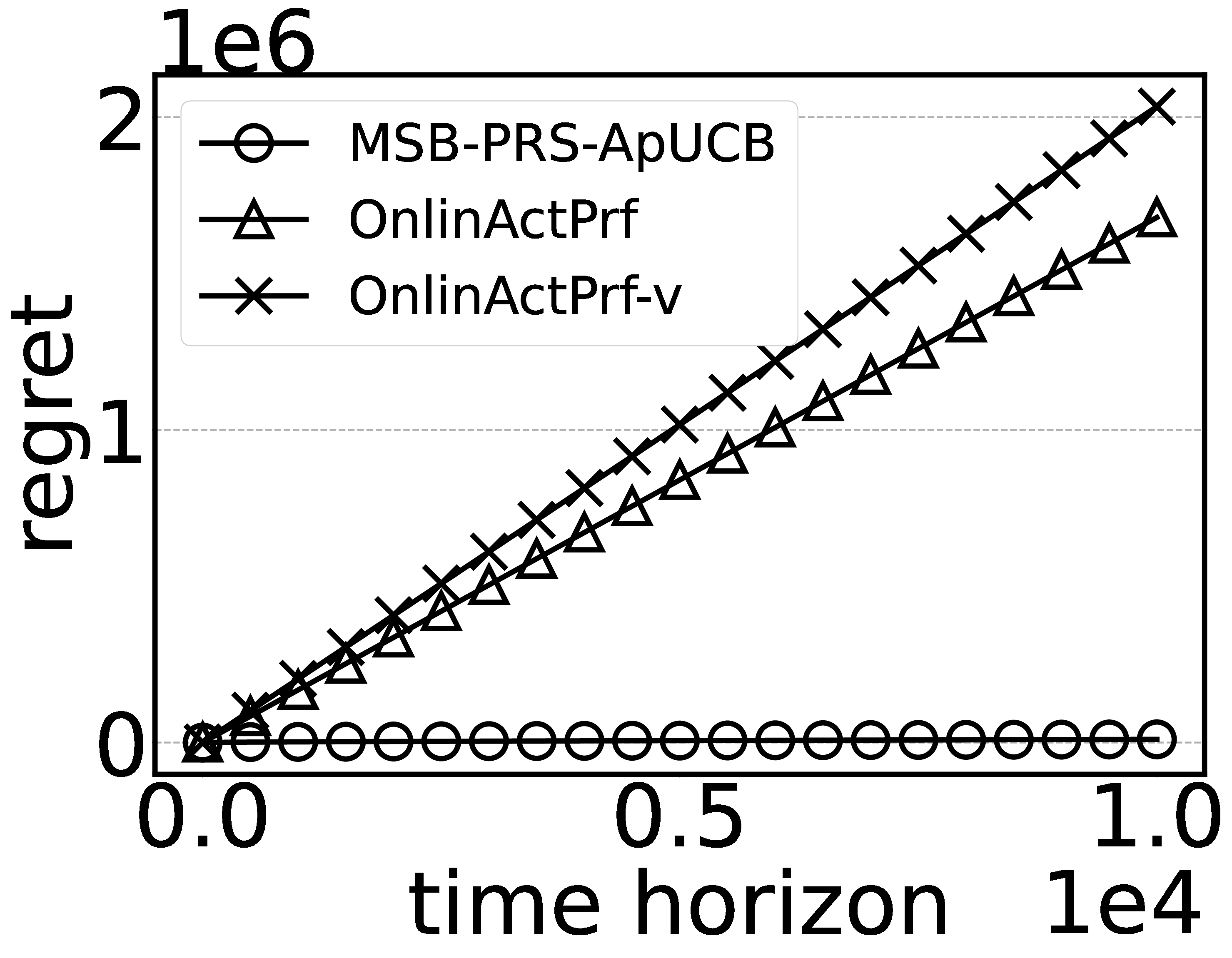}
	}
	\subfloat[$M=15$ \label{fig_narms:3}]
	{
		\includegraphics[width=0.48\linewidth]{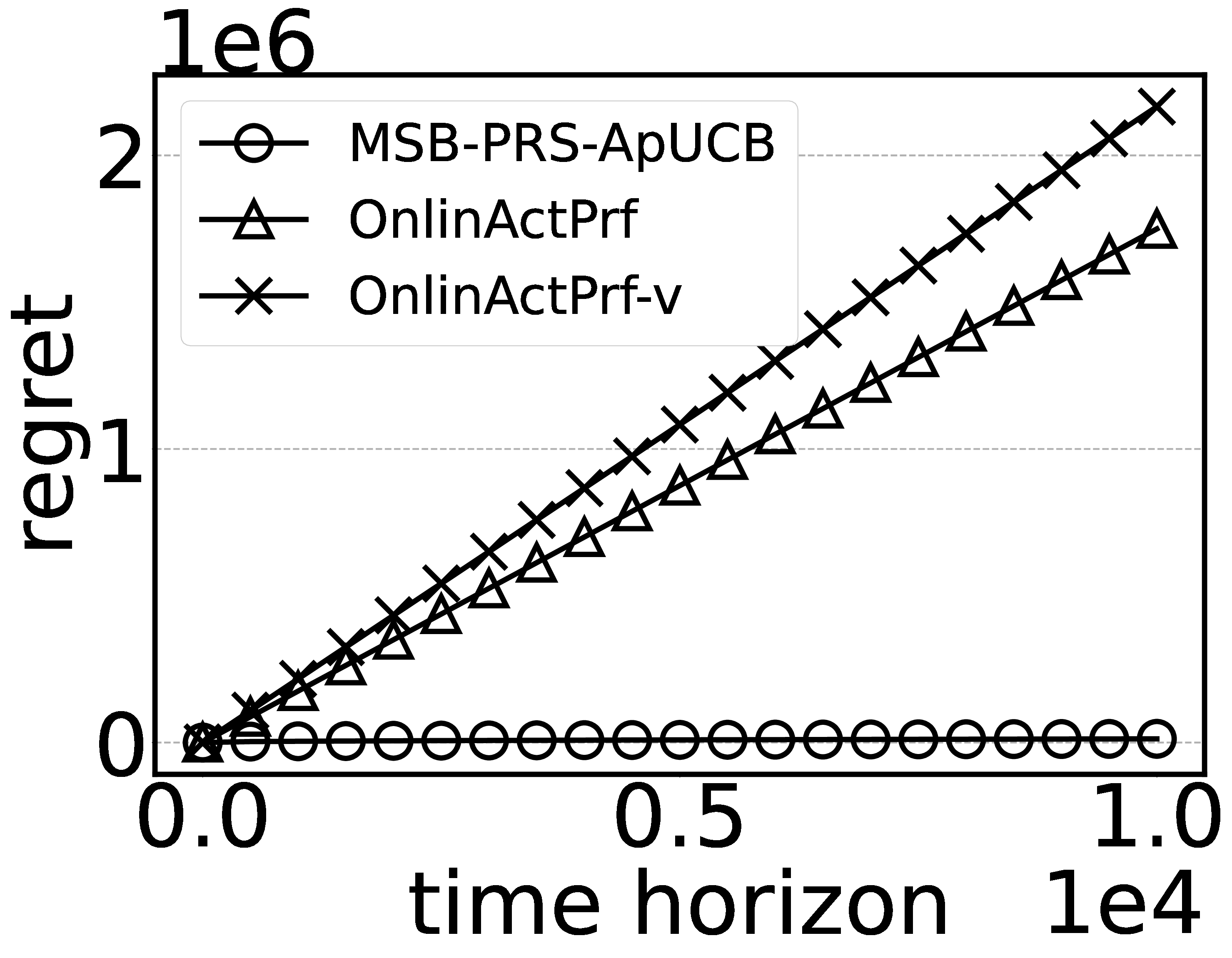}
	}
	\caption{Impact of Number of Arms.}
	\label{fig_narms}
\end{figure}

\subsection{Performance Evaluation} 

{\bf Impact of the number of arms}
We varied the number of arms, denoted as $M$, across three settings: $M = 5$, $10$, and $15$, and plotted the regret of three algorithms. 
In Fig.~\ref{fig_narms:selComp}, it is evident that the regret curves for \texttt{MSB-PRS-ApUCB} under $M = 5$, $10$, 
and $20$ initially exhibit a sharp increase before leveling off, indicating a sub-linear regret. 
Additionally, one can find that the convergence rate of \texttt{MSB-PRS-ApUCB} regret gradually decreases with an increase in $M$.
Fig.~\ref{fig_narms:1} illustrates that the regret curves for \texttt{OnlinActPrf} and \texttt{OnlinActPrf-v} follow a linear trend, while the regret curve for \texttt{MSB-PRS-ApUCB} consistently remains at the bottom. 
This observation confirms that \texttt{MSB-PRS-ApUCB} yields the smallest regret compared to the two baseline algorithms. 
This trend persists even when $M = 10$ and $15$, 
as shown in Fig.~\ref{fig_narms:2} and ~\ref{fig_narms:3}, respectively.   

\begin{figure}[htb]
	\centering
	\subfloat[Regret of \texttt{MSB-PRS-ApUCB} \label{fig_nplays:selComp}]
	{
		\includegraphics[width=0.48\linewidth]{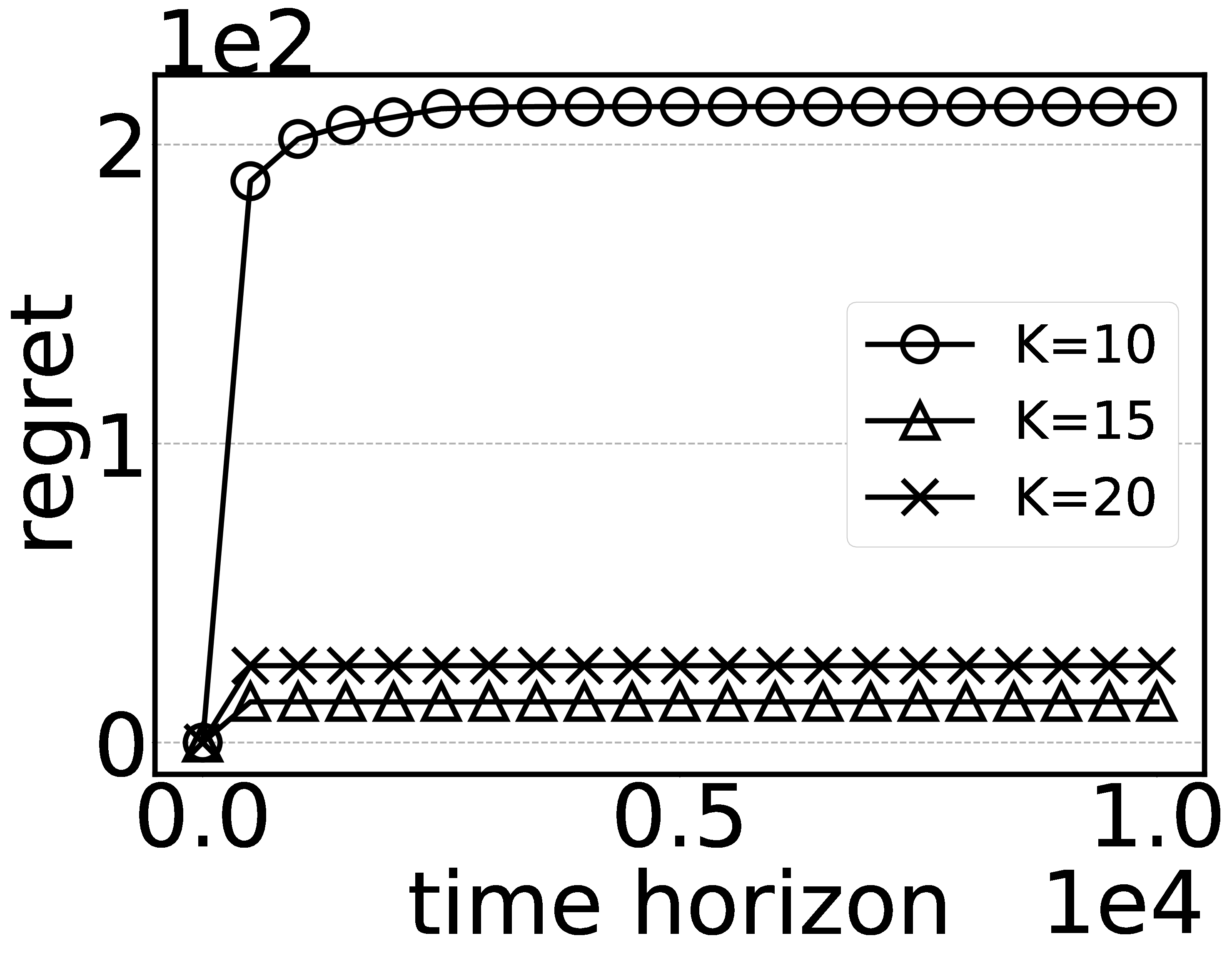}
	}
	\subfloat[ $K=10$ \label{fig_nplays:1}]{
		\includegraphics[width=0.48\linewidth]{pic_eps/algorithmCompare/Arms5_Players_10_Correlation_Middle_Variance_0.1_costFactor_1.eps}
	}\\
	
	\subfloat[$K=15$ \label{fig_nplays:2}]
	{
		\includegraphics[width=0.48\linewidth]{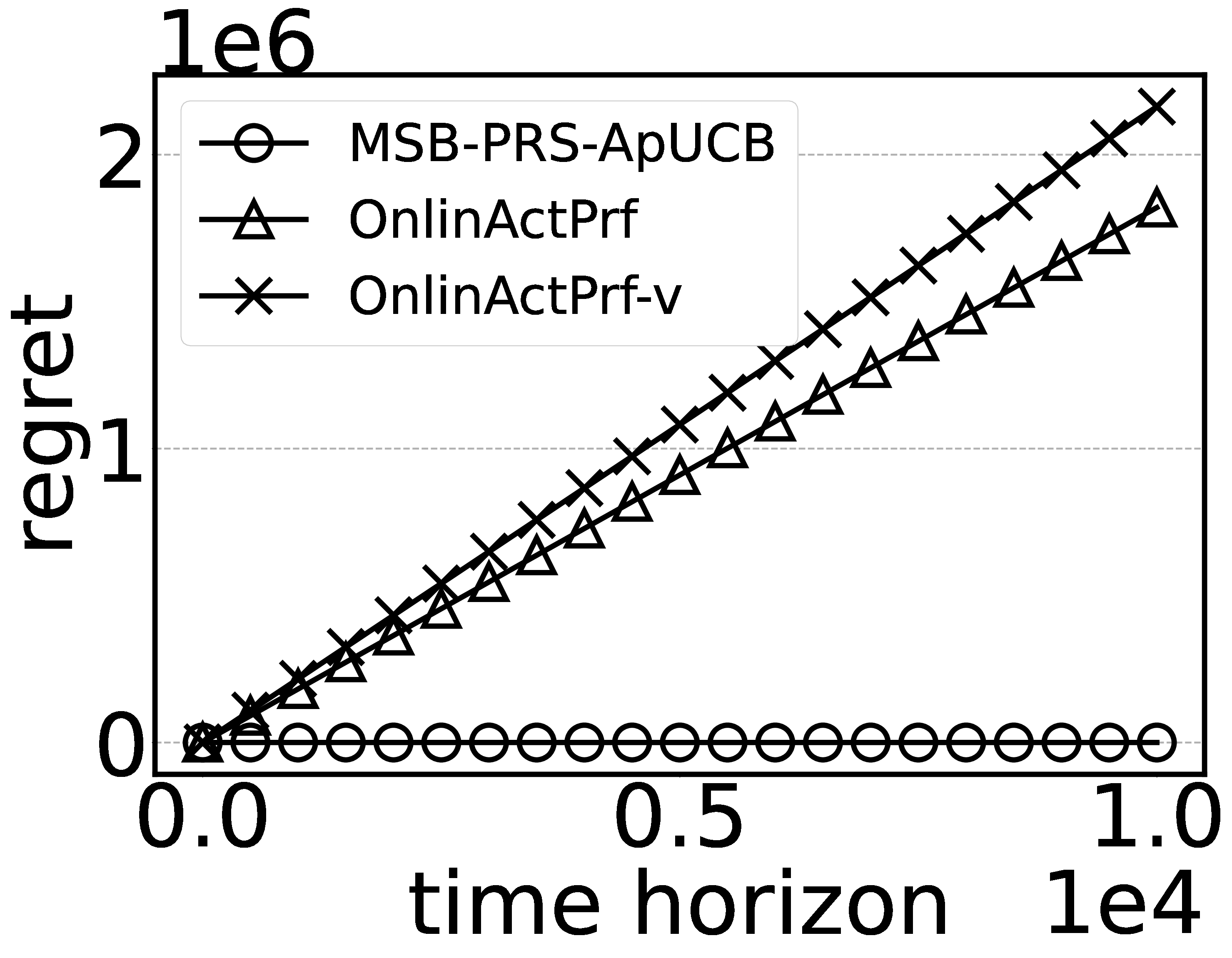}
	}
	\subfloat[$K=20$ \label{fig_nplays:3}]
	{
		\includegraphics[width=0.48\linewidth]{pic_eps/algorithmCompare/Arms5_Players_15_Correlation_Middle_Variance_0.2_costFactor_1.eps}
	}
	\caption{Impact of Number of plays.}
	\label{fig_nplays}
\end{figure}

\noindent
{\bf Impact of the number of plays}
We varied the number of plays, denoted as $K$, across three settings: $K = 10$, $15$, and $20$, and plotted the regret of three algorithms. 
In Fig.~\ref{fig_nplays:selComp}, it is evident that the regret curves for \texttt{MSB-PRS-ApUCB} under $K = 10$, $15$, and $20$ initially exhibit a sharp increase before plateauing, indicating a sub-linear regret. 
Additionally, Fig.~\ref{fig_nplays:1} illustrates that the regret curves for \texttt{OnlinActPrf} and \texttt{OnlinActPrf-v} follow a linear trend, while the regret curve for \texttt{MSB-PRS-ApUCB} consistently remains at the bottom. 
This observation confirms that \texttt{MSB-PRS-ApUCB} yields the smallest regret compared to the two baseline algorithms. 
This trend persists even when $K = 15$ and $20$, as shown in Fig.~\ref{fig_nplays:2} and ~\ref{fig_nplays:3}, respectively.

\begin{figure}[htb]
	\centering
	\subfloat[Regret of \texttt{MSB-PRS-ApUCB} \label{fig_cost:selComp}]
	{
		\includegraphics[width=0.48\linewidth]{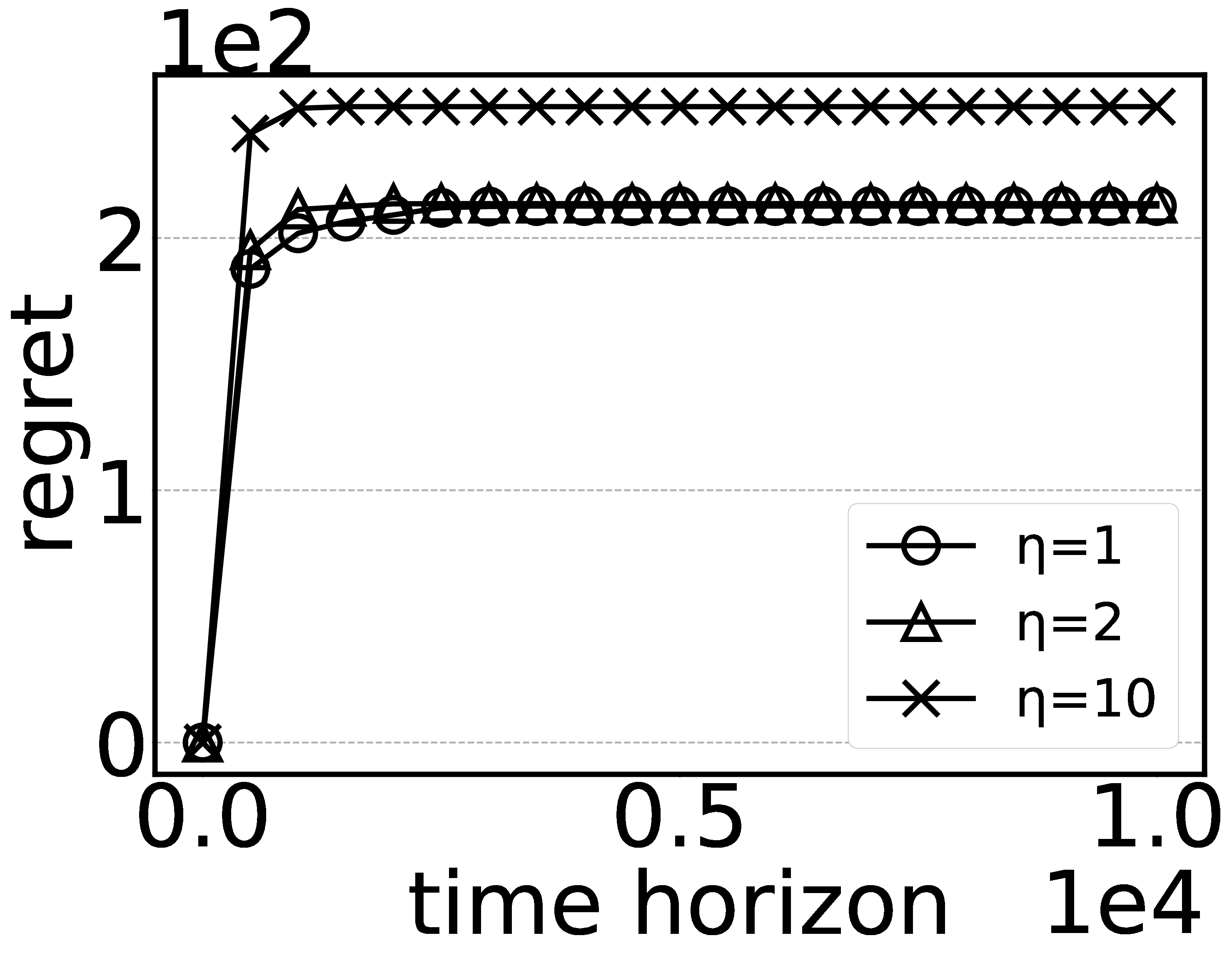}
	}
	\subfloat[ $\eta=1$ \label{fig_cost:1}]{
		\includegraphics[width=0.48\linewidth]{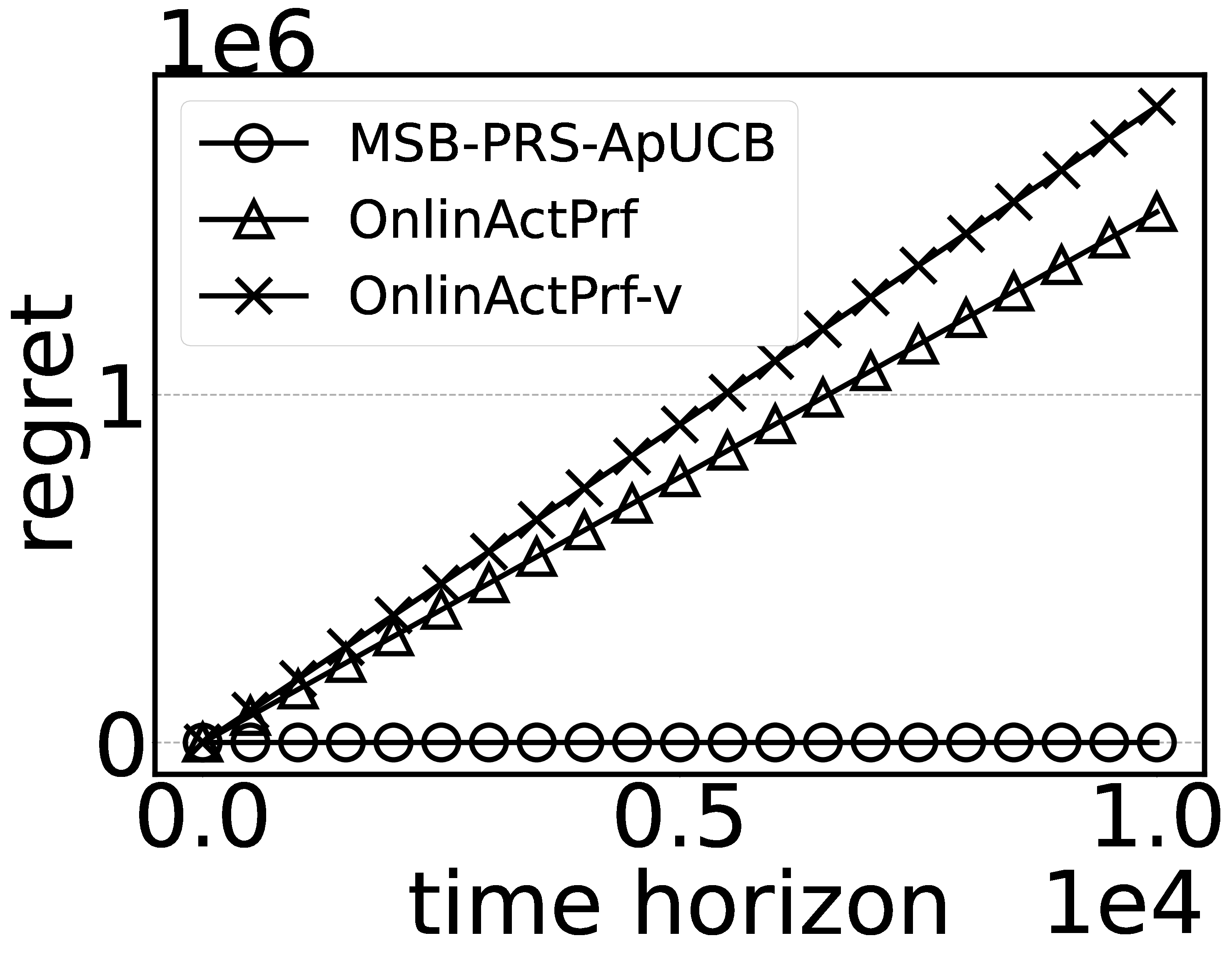}
	}\\
	
	\subfloat[$\eta=2$ \label{fig_cost:2}]
	{
		\includegraphics[width=0.46\linewidth]{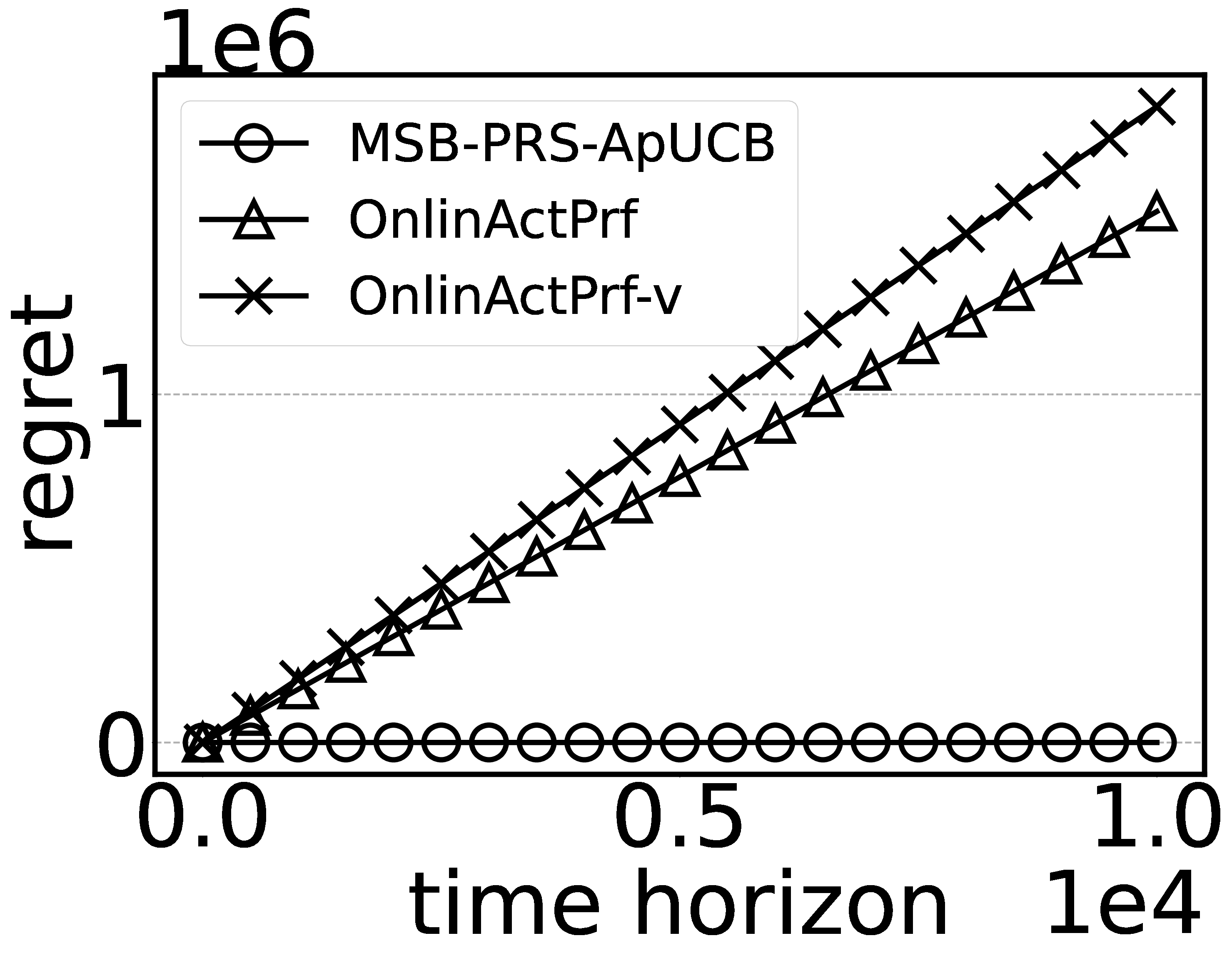}
	}
	\subfloat[$\eta=10$ \label{fig_cost:3}]
	{
		\includegraphics[width=0.48\linewidth]{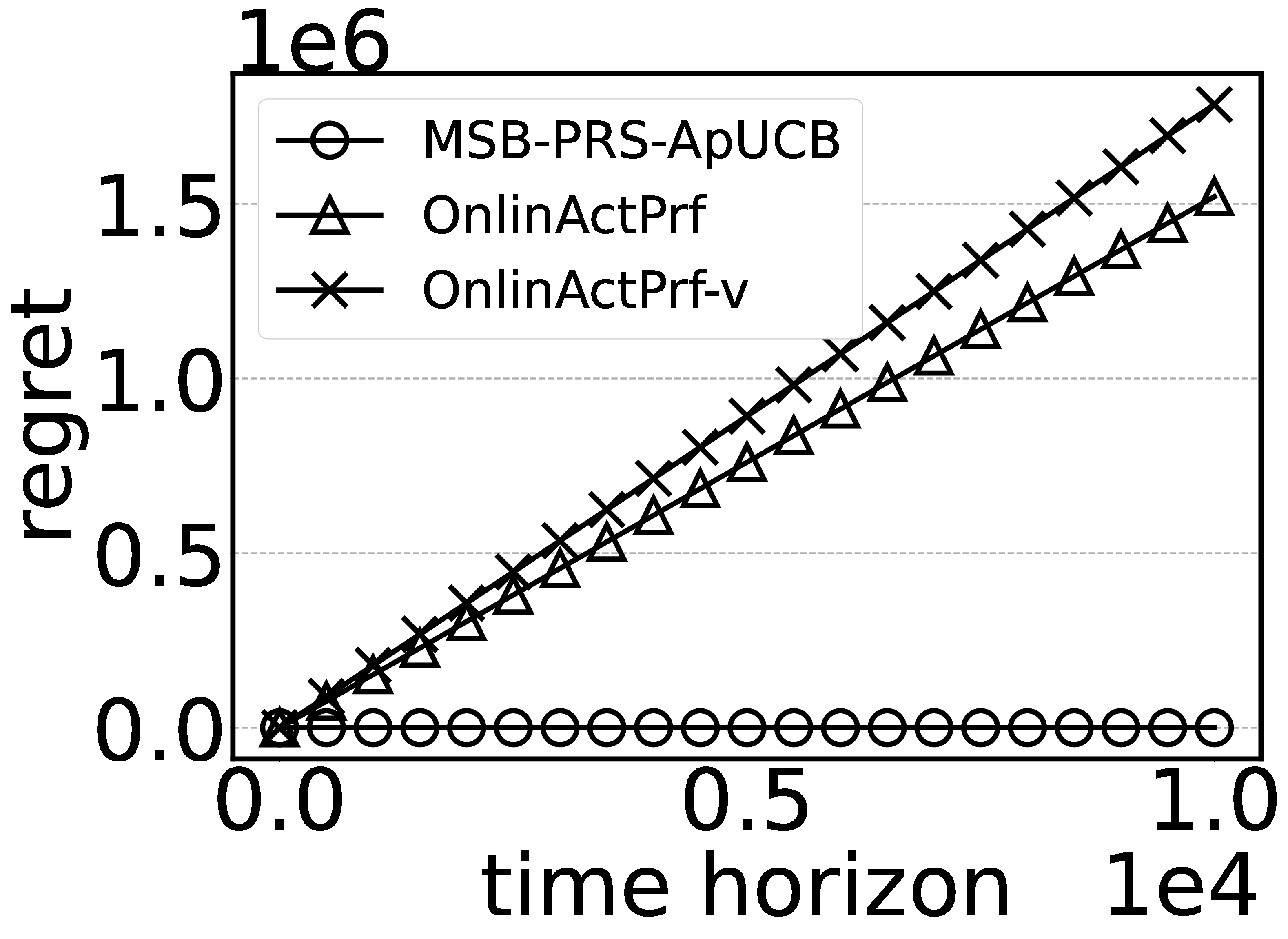}
	}
	\caption{Impact of Movement Cost.}
	\label{fig_cost}
\end{figure}

\noindent
{\bf Impact of movement cost}
We varied the number of plays, denoted as $\eta$, across three settings: $\eta=1$, $2$, and $10$, and plotted the regret of three algorithms. 
In Fig.~\ref{fig_cost:selComp}, it is evident that the regret curves for \texttt{MSB-PRS-ApUCB} under $\eta=1$, $2$, and $10$ initially exhibit a sharp increase before plateauing, indicating a sub-linear regret. 
Additionally, Fig.~\ref{fig_cost:1} illustrates that the regret curves for \texttt{OnlinActPrf} and \texttt{OnlinActPrf-v} follow a linear trend, while the regret curve for \texttt{MSB-PRS-ApUCB} consistently remains at the bottom. 
This observation confirms that \texttt{MSB-PRS-ApUCB} yields the smallest regret compared to the two baseline algorithms. 
This trend persists even when $\eta=2$ and $10$, as shown in Fig.~\ref{fig_cost:2} and ~\ref{fig_cost:3}, respectively.

\begin{figure}[htb]
	\centering
	\subfloat[Regret of \texttt{MSB-PRS-ApUCB} \label{fig_variance:selComp}]
	{
		\includegraphics[width=0.48\linewidth]{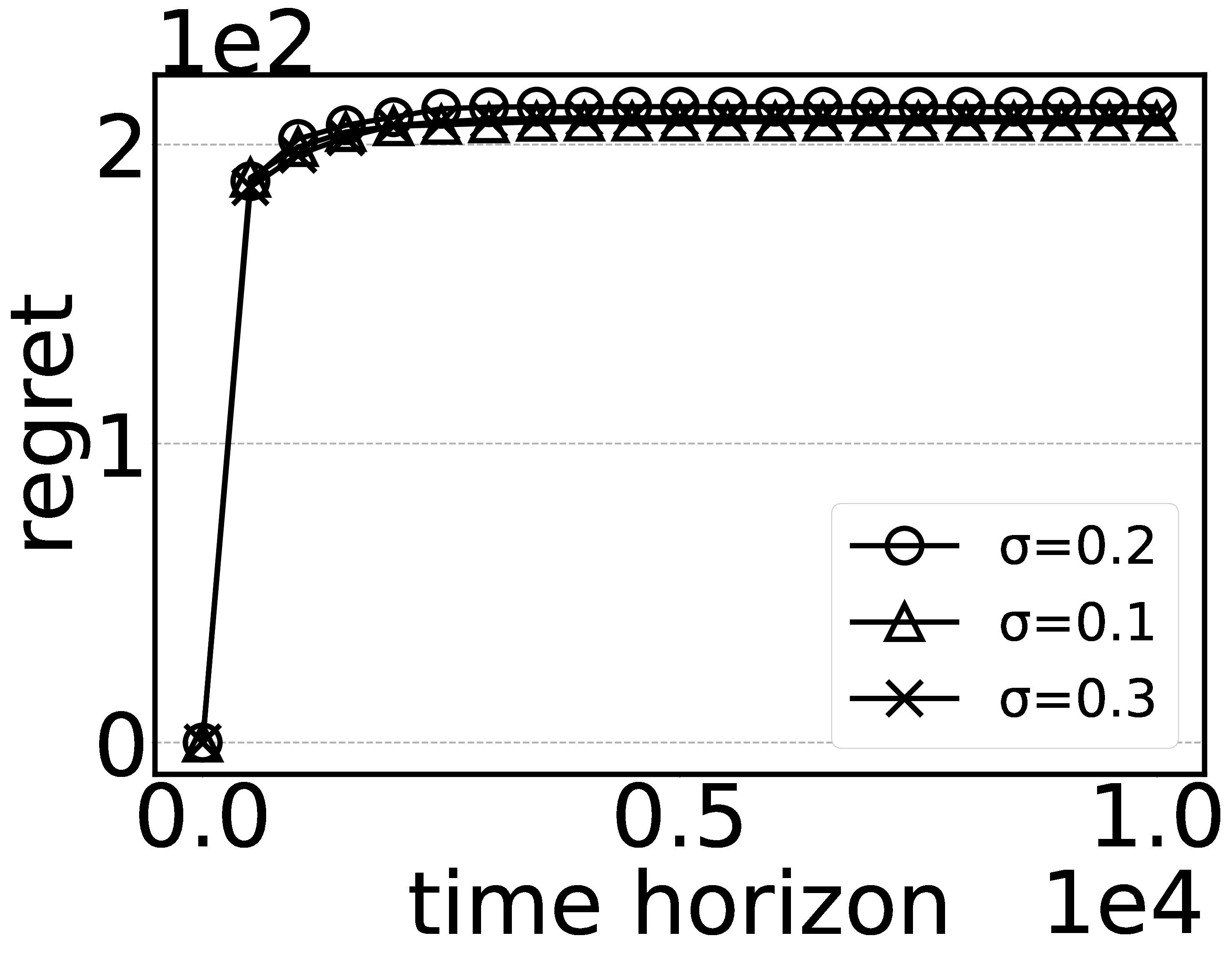}
	}
	\subfloat[ $\sigma=0.1$ \label{fig_variance:1}]{
		\includegraphics[width=0.483\linewidth]{pic_eps/algorithmCompare/Arms5_Players_10_Correlation_Middle_Variance_0.1_costFactor_1.eps}
	}\\
	\subfloat[$\sigma=0.2$ \label{fig_variance:2}]
	{
		\includegraphics[width=0.48\linewidth]{pic_eps/algorithmCompare/Arms5_Players_10_Correlation_Middle_Variance_0.2_costFactor_1.eps}
	}
	\subfloat[$\sigma=0.3$ \label{fig_variance:3}]
	{
		\includegraphics[width=0.48\linewidth]{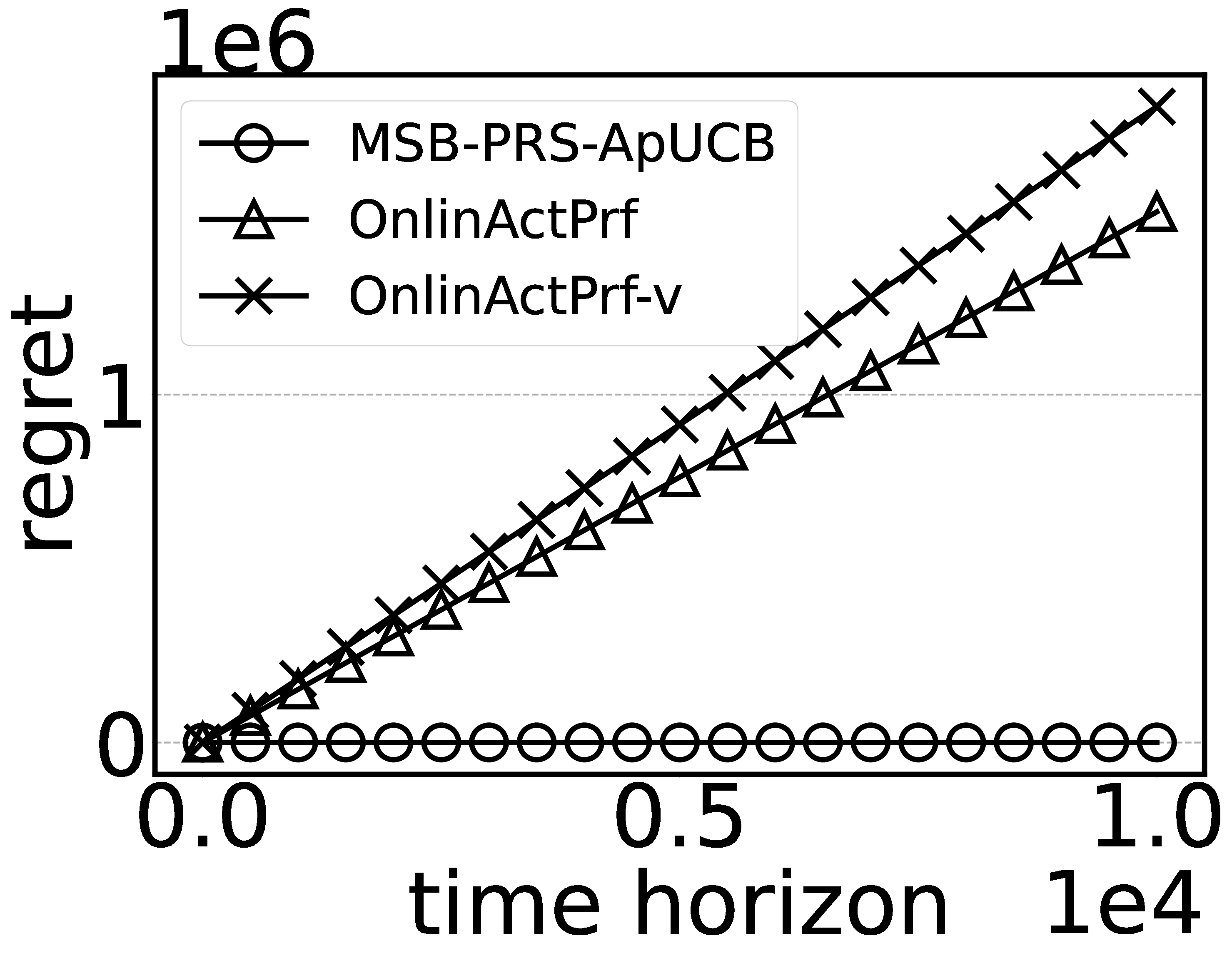}
	}
	\caption{Impact of Standard Deviation of Reward.}
	\label{fig_variance}
\end{figure}

\noindent
{\bf Impact of the standard deviation of reward}
We varied the standard deviation of reward, denoted as $\sigma$, 
across three settings: $\sigma = 0.1$, $0.2$, and $0.3$, and plotted the regret of three algorithms. 
In Fig.~\ref{fig_variance:selComp}, it is evident that the regret curves for \texttt{MSB-PRS-ApUCB} under $\sigma = 0.1$, $0.2$, and $0.3$ initially exhibit a sharp increase before plateauing, indicating a sub-linear regret. 
Additionally, Fig.~\ref{fig_variance:1} illustrates that the regret curves for \texttt{OnlinActPrf} and \texttt{OnlinActPrf-v} follow a linear trend, while the regret curve for \texttt{MSB-PRS-ApUCB} consistently remains at the bottom. 
This observation confirms that \texttt{MSB-PRS-ApUCB} yields the smallest regret compared to the two baseline algorithms. 
This trend persists even when $\sigma = 0.2$ and $0.3$, as shown in Fig.~\ref{fig_variance:2} and ~\ref{fig_variance:3}, respectively.

\section{Conclusion}

This paper proposes  MSB-PRS.  
An algorithm is designed to locate the optimal play allocation policy with a complexity of 
$O(M^3K^3)$.   
Instance independent and instance dependent regret lower bounds of 
$\Omega( \alpha_1 \sigma \sqrt{KM T} )$ and 
$\Omega(\alpha_1 \sigma^2 \frac{M}{\Delta} \ln T)$ 
are proved respectively.  
An approximate UCB based algorithm is designed 
which has a per round 
computational complexity of $O(M^3K^3)$ 
and has sublinear independent and dependent regret upper bounds 
matching the corresponding 
lower bounds up to acceptable factors.    

\section*{Acknowledgments}

This work was supported by the National Natural Science Foundation of China (No. 62476261).   
Tao Tan is the corresponding author. 

\clearpage

\bibliography{reference}

@article{7738348,
  author={Crouse, David F.},
  journal={IEEE Transactions on Aerospace and Electronic Systems}, 
  title={On implementing 2D rectangular assignment algorithms}, 
  year={2016},
  volume={52},
  number={4},
  pages={1679-1696},
  doi={10.1109/TAES.2016.140952}}

@article{Agrawal1990,
  title={Multi-armed bandit problems with multiple plays and switching cost},
  author={Agrawal, R and Hegde, M and Teneketzis, D and others},
  journal={Stochastics and Stochastic reports},
  volume={29},
  number={4},
  pages={437--459},
  year={1990}
}

@article{anandkumar2011distributed,
  title     = {Distributed algorithms for learning and cognitive medium access with logarithmic regret},
  author    = {Anandkumar, Animashree and Michael, Nithin and Tang, Ao Kevin and Swami, Ananthram},
  journal   = {IEEE Journal on Selected Areas in Communications},
  volume    = {29},
  number    = {4},
  pages     = {731--745},
  year      = {2011},
  publisher = {IEEE}
}

@article{Agarwal2022multi,
  title={Multi-agent multi-armed bandits with limited communication},
  author={Agarwal, Mridul and Aggarwal, Vaneet and Azizzadenesheli, Kamyar},
  journal={The Journal of Machine Learning Research},
  volume={23},
  number={1},
  pages={9529--9552},
  year={2022},
  publisher={JMLRORG}
}

@article{anantharam1987asymptotically,
  title     = {Asymptotically efficient allocation rules for the multiarmed bandit problem with multiple plays-part i: Iid rewards},
  author    = {Anantharam, Venkatachalam and Varaiya, Pravin and Walrand, Jean},
  journal   = {IEEE Transactions on Automatic Control},
  volume    = {32},
  number    = {11},
  pages     = {968--976},
  year      = {1987},
  publisher = {IEEE}
}

@article{anantharam1987b,
  title={Asymptotically efficient allocation rules for the multiarmed bandit problem with multiple plays-Part II: Markovian rewards},
  author={Anantharam, Venkatachalam and Varaiya, Pravin and Walrand, Jean},
  journal={IEEE Transactions on Automatic Control},
  volume={32},
  number={11},
  pages={977--982},
  year={1987},
  publisher={IEEE}
}

@article{bistritz2018distributed,
  title   = {Distributed multi-player bandits-a game of thrones approach},
  author  = {Bistritz, Ilai and Leshem, Amir},
  journal = {Advances in Neural Information Processing Systems (NeurIPS)},
  year    = {2018}
}

@article{cesa2012combinatorial,
  title     = {Combinatorial bandits},
  author    = {Cesa-Bianchi, Nicolo and Lugosi, G{\'a}bor},
  journal   = {Journal of Computer and System Sciences},
  volume    = {78},
  number    = {5},
  pages     = {1404--1422},
  year      = {2012},
  publisher = {Elsevier}
}

@inproceedings{chen2013combinatorial,
  title        = {Combinatorial multi-armed bandit: General framework and applications},
  author       = {Chen, Wei and Wang, Yajun and Yuan, Yang},
  booktitle    = {International Conference on Machine Learning},
  pages        = {151--159},
  year         = {2013},
  organization = {PMLR}
}

@article{chen2016combinatorial,
  title     = {Combinatorial multi-armed bandit and its extension to probabilistically triggered arms},
  author    = {Chen, Wei and Wang, Yajun and Yuan, Yang and Wang, Qinshi},
  journal   = {The Journal of Machine Learning Research},
  volume    = {17},
  number    = {1},
  pages     = {1746--1778},
  year      = {2016},
  publisher = {JMLR. org}
}

@article{chen2021distributed,
  title={Distributed learning dynamics of multi-armed bandits for edge intelligence},
  author={Chen, Shuzhen and Tao, Youming and Yu, Dongxiao and Li, Feng and Gong, Bei},
  journal={Journal of Systems Architecture},
  volume={114},
  pages={101919},
  year={2021},
  publisher={Elsevier}
}

@inproceedings{Chen2022,
  title={An Online Learning Approach to Sequential User-Centric Selection Problems},
  author={Chen, Junpu and Xie, Hong},
  booktitle={Proceedings of the AAAI Conference on Artificial Intelligence},
  volume={36},
  number={6},
  pages={6231--6238},
  year={2022}
}

@inproceedings{Combes2015,
  title={Combinatorial Bandits Revisited},
  author={Combes, Richard and Talebi, Sadegh and Prouti{\`e}re, Alexandre and Lelarge, Marc},
  booktitle={NIPS 2015-Twenty-ninth Conference on Neural Information Processing Systems},
  year={2015}
}

@inproceedings{combes2015learning,
  title     = {Learning to rank: Regret lower bounds and efficient algorithms},
  author    = {Combes, Richard and Magureanu, Stefan and Proutiere, Alexandre and Laroche, Cyrille},
  booktitle = {Proceedings of the 2015 ACM SIGMETRICS International Conference on Measurement and Modeling of Computer Systems},
  pages     = {231--244},
  year      = {2015}
}

@article{gai2012combinatorial,
  author  = {Y. {Gai} and B. {Krishnamachari} and R. {Jain}},
  journal = {IEEE/ACM Transactions on Networking},
  title   = {Combinatorial Network Optimization With Unknown Variables: Multi-Armed Bandits With Linear Rewards and Individual Observations},
  year    = {2012},
  volume  = {20},
  number  = {5},
  pages   = {1466-1478},
  doi     = {10.1109/TNET.2011.2181864}
}

@article{gao2022combination,
  title={Combination of auction theory and multi-armed bandits: Model, algorithm, and application},
  author={Gao, Guoju and Huang, Sijie and Huang, He and Xiao, Mingjun and Wu, Jie and Sun, Yu-E and Zhang, Sheng},
  journal={IEEE Transactions on Mobile Computing},
  year={2022},
  publisher={IEEE}
}

@article{Jun2004,
  title={A survey on the bandit problem with switching costs},
  author={Jun, Tackseung},
  journal={de Economist},
  volume={152},
  number={4},
  pages={513--541},
  year={2004},
  publisher={Springer}
}

@inproceedings{komiyama_optimal_2015,
  title        = {Optimal regret analysis of {T}hompson sampling in stochastic multi-armed bandit problem with multiple plays},
  author       = {Komiyama, Junpei and Honda, Junya and Nakagawa, Hiroshi},
  booktitle    = {International Conference on Machine Learning},
  pages        = {1152--1161},
  year         = {2015},
  organization = {PMLR}
}

@inproceedings{komiyama2017position,
  title     = {Position-based multiple-play bandit problem with unknown position bias},
  author    = {Komiyama, Junpei and Honda, Junya and Takeda, Akiko},
  booktitle = {Proceedings of the 31st International Conference on Neural Information Processing Systems},
  pages     = {5005--5015},
  year      = {2017}
}

@inproceedings{kveton2014matroid,
  title     = {Matroid bandits: fast combinatorial optimization with learning},
  author    = {Kveton, Branislav and Wen, Zheng and Ashkan, Azin and Eydgahi, Hoda and Eriksson, Brian},
  booktitle = {Proceedings of the Thirtieth Conference on Uncertainty in Artificial Intelligence},
  pages     = {420--429},
  year      = {2014}
}

@inproceedings{kveton2015combinatorial,
  title     = {Combinatorial cascading bandits},
  author    = {Kveton, Branislav and Wen, Zheng and Ashkan, Azin and Szepesv{\'a}ri, Csaba},
  booktitle = {Proceedings of the 28th International Conference on Neural Information Processing Systems-Volume 1},
  pages     = {1450--1458},
  year      = {2015}
}

@inproceedings{Kveton2015,
  title={Tight regret bounds for stochastic combinatorial semi-bandits},
  author={Kveton, Branislav and Wen, Zheng and Ashkan, Azin and Szepesvari, Csaba},
  booktitle={Artificial Intelligence and Statistics},
  pages={535--543},
  year={2015},
  organization={PMLR}
}

@inproceedings{lagree2016multiple,
  title     = {Multiple-play bandits in the position-based model},
  author    = {Lagr{\'e}e, Paul and Vernade, Claire and Capp{\'e}, Olivier},
  booktitle = {Proceedings of the 30th International Conference on Neural Information Processing Systems},
  pages     = {1605--1613},
  year      = {2016}
}

@book{lattimore2020bandit,
  title     = {Bandit algorithms},
  author    = {Lattimore, Tor and Szepesv{\'a}ri, Csaba},
  year      = {2020},
  publisher = {Cambridge University Press}
}

@article{Lesage2017,
  title={The multi-armed bandit with stochastic plays},
  author={Lesage-Landry, Antoine and Taylor, Joshua A},
  journal={IEEE Transactions on Automatic Control},
  volume={63},
  number={7},
  pages={2280--2286},
  year={2017},
  publisher={IEEE}
}

@article{Luedtke2019,
  title={Asymptotically optimal algorithms for budgeted multiple play bandits},
  author={Luedtke, Alex and Kaufmann, Emilie and Chambaz, Antoine},
  journal={Machine Learning},
  volume={108},
  pages={1919--1949},
  year={2019},
  publisher={Springer}
}

@article{Maillard2017,
  title={Basic Concentration Properties of Real-Valued Distributions},
  author={Maillard, Odalric-Ambrym},
  year={2017}
}

@article{Moulos2020,
  title={Finite-time analysis of round-robin kullback-leibler upper confidence bounds for optimal adaptive allocation with multiple plays and Markovian rewards},
  author={Moulos, Vrettos},
  journal={Advances in Neural Information Processing Systems},
  volume={33},
  pages={7863--7874},
  year={2020}
}

@ARTICLE{Ouyang2023,
  author={Ouyang, Tao and Chen, Xu and Zhou, Zhi and Li, Rui and Tang, Xin},
  journal={IEEE Transactions on Mobile Computing}, 
  title={Adaptive User-Managed Service Placement for Mobile Edge Computing via Contextual Multi-Armed Bandit Learning}, 
  year={2023},
  volume={22},
  number={3},
  pages={1313-1326},
  doi={10.1109/TMC.2021.3106746}}

@inproceedings{Ouyang2019,
  title={Adaptive user-managed service placement for mobile edge computing: An online learning approach},
  author={Ouyang, Tao and Li, Rui and Chen, Xu and Zhou, Zhi and Tang, Xin},
  booktitle={IEEE INFOCOM 2019-IEEE conference on computer communications},
  pages={1468--1476},
  year={2019},
  organization={IEEE}
}

@inproceedings{rosenski2016multi,
  title        = {Multi-player bandits--a musical chairs approach},
  author       = {Rosenski, Jonathan and Shamir, Ohad and Szlak, Liran},
  booktitle    = {International Conference on Machine Learning},
  pages        = {155--163},
  year         = {2016},
  organization = {PMLR}
}

@inproceedings{wang2020optimal,
  title        = {Optimal algorithms for multiplayer multi-armed bandits},
  author       = {Wang, Po-An and Proutiere, Alexandre and Ariu, Kaito and Jedra, Yassir and Russo, Alessio},
  booktitle    = {International Conference on Artificial Intelligence and Statistics},
  pages        = {4120--4129},
  year         = {2020},
  organization = {PMLR}
}

@inproceedings{Wang2022,
  author       = {Xuchuang Wang and
                  Hong Xie and
                  John C. S. Lui},
  editor       = {Kamalika Chaudhuri and
                  Stefanie Jegelka and
                  Le Song and
                  Csaba Szepesv{\'{a}}ri and
                  Gang Niu and
                  Sivan Sabato},
  title        = {Multiple-Play Stochastic Bandits with Shareable Finite-Capacity Arms},
  booktitle    = {International Conference on Machine Learning, {ICML} 2022, 17-23 July
                  2022, Baltimore, Maryland, {USA}},
  series       = {Proceedings of Machine Learning Research},
  volume       = {162},
  pages        = {23181--23212},
  publisher    = {{PMLR}},
  year         = {2022},
  url          = {https://proceedings.mlr.press/v162/wang22af.html},
  bibsource    = {dblp computer science bibliography, https://dblp.org}
}

@inproceedings{Wang2022a,
  author       = {Xuchuang Wang and
                  Hong Xie and
                  John C. S. Lui},
  editor       = {Luc De Raedt},
  title        = {Multi-Player Multi-Armed Bandits with Finite Shareable Resources Arms:
                  Learning Algorithms {\&} Applications},
  booktitle    = {Proceedings of the Thirty-First International Joint Conference on
                  Artificial Intelligence, {IJCAI} 2022, Vienna, Austria, 23-29 July
                  2022},
  pages        = {3537--3543},
  publisher    = {ijcai.org},
  year         = {2022},
  url          = {https://doi.org/10.24963/ijcai.2022/491},
  doi          = {10.24963/IJCAI.2022/491},
  bibsource    = {dblp computer science bibliography, https://dblp.org}
}

@inproceedings{wen2017online,
  title     = {Online influence maximization under independent cascade model with semi-bandit feedback},
  author    = {Wen, Zheng and Kveton, Branislav and Valko, Michal and Vaswani, Sharan},
  booktitle = {Neural Information Processing Systems},
  pages     = {1--24},
  year      = {2017}
}

@inproceedings{Xia2016,
  title={Budgeted Multi-Armed Bandits with Multiple Plays.},
  author={Xia, Yingce and Qin, Tao and Ma, Weidong and Yu, Nenghai and Liu, Tie-Yan},
  booktitle={IJCAI},
  pages={2210--2216},
  year={2016}
}

@inproceedings{Xu2023,
author = {Xu, Renzhe and Wang, Haotian and Zhang, Xingxuan and Li, Bo and Cui, Peng},
title = {Competing for shareable arms in multi-player multi-armed bandits},
year = {2023},
publisher = {JMLR.org},
booktitle = {Proceedings of the 40th International Conference on Machine Learning},
articleno = {1612},
numpages = {33},
location = {Honolulu, Hawaii, USA},
series = {ICML'23}
}

@inproceedings{Yuan2023,
author = {Yuan, Jianjun and Woon, Wei Lee and Coba, Ludovik},
title = {Adversarial Sleeping Bandit Problems with Multiple Plays: Algorithm and Ranking Application},
year = {2023},
isbn = {9798400702419},
publisher = {Association for Computing Machinery},
address = {New York, NY, USA},
url = {https://doi.org/10.1145/3604915.3608824},
doi = {10.1145/3604915.3608824},
booktitle = {Proceedings of the 17th ACM Conference on Recommender Systems},
pages = {744–749},
numpages = {6},
location = {Singapore, Singapore},
series = {RecSys '23}
}

@inproceedings{Zhou2018,
  title={Budget-constrained multi-armed bandits with multiple plays},
  author={Zhou, Datong and Tomlin, Claire},
  booktitle={Proceedings of the AAAI Conference on Artificial Intelligence},
  volume={32},
  number={1},
  year={2018}
}


\clearpage

 \onecolumn

\subsection{Proof of Lemma \ref{lem:offline:ActionToMatching}}

We first prove the $\mathcal{U}$-saturated property.  
Note that each action profile $\boldsymbol{a}$ assigns all plays to arms 
and each arm is assigned to only one arm. Thus, it holds that 
$
\{
u | (u,v) \in \widetilde{\mathcal{M}}(\boldsymbol{a})
\} 
= \mathcal{U}.  
$

Now we prove the $\mathcal{V}$-monotone and priority compatible property.  
Note that all the plays assigned to an arm are ordered 
based on $\ell_k (\boldsymbol{a})$.  
This order list is monotone and priority compatible.  
Thus, $\widetilde{\mathcal{M}}(\boldsymbol{a})$ is monotone and 
priority compatible.

We prove the utility preserving property.  
Let
$
\mathcal{J}_m 
= \{
j | a_j = m
\}
$ 
denote the set of all plays that pull arm $m$.  
By the monotone and priority compatible property of 
$\ell_k (\boldsymbol{a})$ and some basic arguments we have: 
\begin{align*} 
U(\boldsymbol{a}; \boldsymbol{\mu}, \boldsymbol{P})  
& 
=
\sum_{m\in[M]} U_{m}(\boldsymbol{a}; \mu_m, \boldsymbol{P}_m)  
\\
& 
=
\sum_{m\in[M]}
\sum_{k \in [K]}
\boldsymbol{1}_{\{a_k = m\}}  
\Lambda_m (k, \ell_k (\boldsymbol{a}))
\\
& 
=\sum_{m\in[M]}
\sum_{k \in [K]}
\boldsymbol{1}_{\{a_k = m\}}  
W (u_k, v_{m,\ell_k (\boldsymbol{a})})
\\
&
=\sum_{m\in[M]}
\sum_{k \in [K]}
\boldsymbol{1}_{\{\phi_{m,\ell_k (\boldsymbol{a}) }= k\}}  
W (u_k, v_{m,\ell_k (\boldsymbol{a})})
\\
& 
=\sum_{m\in[M]}
\sum_{k \in [K]} 
\sum_{j=1}^{|\mathcal{J}_m|}
\boldsymbol{1}_{\{\phi_{m,j}= k\}}  
W (u_k, v_{m,j})
\\
& 
=\sum_{m\in[M]}
\sum_{j=1}^{|\mathcal{J}_m|}
\sum_{k \in [K]} 
\boldsymbol{1}_{\{\phi_{m,j}= k\}}  
W (u_k, v_{m,j})
\\
& 
=\sum_{m\in[M]}
\sum_{j=1}^{|\mathcal{J}_m|}
W (u_k, v_{m,j})
= 
 \sum_{ (u,v) \in \widetilde{\mathcal{M}}(\boldsymbol{a}) } 
W(u,v).  
\end{align*}

Finally, we prove the uniqueness.  
Consider two action profiles $\boldsymbol{a}$ and $\boldsymbol{a}'$ 
such that $\boldsymbol{a} \neq \boldsymbol{a}'$.   
Then, we know that there exists at least one 
element that $\boldsymbol{a}$ and $\boldsymbol{a}'$ do not agree.  
This means that there exists at least one play 
that pulls different arms under $\boldsymbol{a}$ and $\boldsymbol{a}'$ respectively.  
Then, we conclude the uniqueness.  
This proof is then complete.

\subsection{Proof of Lemma \ref{lem:offline:MatchingToAction}}

One can easily verify that $\widetilde{a}_k (\mathcal{M})$ 
is the arm pulled by play $k$ under the matching $\mathcal{M}$.  
Since the matching $\mathcal{M}$ is  $\mathcal{U}$-saturated, 
each play is assigned to one arm.  
Namely, $\widetilde{\boldsymbol{a}} (\mathcal{M})$ forms a valid assignment.  
The remaining thing is to show the the utility of 
$\widetilde{\boldsymbol{a}} (\mathcal{M})$ matches the 
sum of weights of $\mathcal{M}$.  
Let $\widetilde{\mathcal{V}}_m$ denote the end points of $\mathcal{M}$ 
on the arm node side 
that belong to the set $\mathcal{V}_m$, 
formally
\[
\widetilde{\mathcal{V}}_m 
= 
\{ v | (u,v) \in \mathcal{M} \}
\cap \mathcal{V}_m.  
\]
Note that the matching $\mathcal{M}$ is $\mathcal{V}$-monotone.  
This means that $\widetilde{\mathcal{V}}_m$ can be expressed as 
$
\widetilde{\mathcal{V}}_m 
= \{ v_{m,1}, \ldots, 
v_{m,|\widetilde{\mathcal{V}}_m|} 
\}
$.  
Furthermore, $(u_{\phi_{m,j} (\mathcal{M})}, v_{m,j})$ is an edge of 
the matching with the arm side end point in the set 
$\widetilde{\mathcal{V}}_m$.  
The total weights of the matching $\mathcal{M}$ can be decomposed as 
\[
\sum_{(u,v)\in  \mathcal{M}} W(u,v)
=
\sum_{m \in [M]}
\sum_{j=1}^{|\widetilde{\mathcal{V}}_m|} 
W (u_{\phi_{m,j} (\mathcal{M})}, 
v_{m,j}) 
=
\sum_{m \in [M]}
\sum_{j=1}^{|\widetilde{\mathcal{V}}_m|} \Lambda_m (\phi_{m,j} (\mathcal{M}), 
j).  
\]
We next complete this proof by showing that 
\[ 
\sum_{j=1}^{|\widetilde{\mathcal{V}}_m|} 
\Lambda_m (\phi_{m,j} (\mathcal{M}), j)
=
U_{m}(\widetilde{\boldsymbol{a}} (\mathcal{M}); \mu_m, \boldsymbol{P}_m).  
\]
The matching $\mathcal{M}$ being priority compatible implies that 
\[
\ell_{\phi_{m,j} (\mathcal{M})} (\widetilde{\boldsymbol{a}} (\mathcal{M})) = j. 
\]
Then, it follows that: 
\begin{align*}
\sum_{j=1}^{|\widetilde{\mathcal{V}}_m|} 
\Lambda_m (\phi_{m,j} (\mathcal{M}), j)
&
=
\sum_{j=1}^{|\widetilde{\mathcal{V}}_m|} \Lambda_m (\phi_{m,j} (\mathcal{M}), 
\ell_{\phi_{m,j} (\mathcal{M})} (\widetilde{\boldsymbol{a}} (\mathcal{M})))
\\
& 
=
\sum_{j=1}^{|\widetilde{\mathcal{V}}_m|} 
\sum_{k \in [K] } 
\boldsymbol{1}_{ \{ \phi_{m,j} (\mathcal{M}) = k \}} 
\Lambda_m (k, 
\ell_{k} (\widetilde{\boldsymbol{a}} (\mathcal{M})))
\\
& 
=
\sum_{k \in [K] } 
\sum_{j=1}^{|\widetilde{\mathcal{V}}_m|} 
\boldsymbol{1}_{ \{ \phi_{m,j} (\mathcal{M}) = k \}} 
\Lambda_m (k, 
\ell_{k} (\widetilde{\boldsymbol{a}} (\mathcal{M})))
\\
& 
=
\sum_{k \in [K] }  
\boldsymbol{1}_{ \{ \widetilde{a}_k (\mathcal{M}) =m \}} 
\Lambda_m (k, 
\ell_{k} (\widetilde{\boldsymbol{a}} (\mathcal{M})))
\\
&  
= 
U_{m}(\widetilde{\boldsymbol{a}} (\mathcal{M}); \mu_m, \boldsymbol{P}_m), 
\end{align*}
This proof is then complete.

\subsection{Proof of Lemma \ref{lem:online:mu}}

Note that the range of $\boldsymbol{1}_{\{D^{(s)}_{m} \geq d\}}$ is  within $\{0,1\}$.  
It is a $\frac{1}{2}$-subgaussian random variable.  
Note that $D^{(t)}_m$'s are independent across $t$ and $m$.  
Lemma 10 of \cite{Maillard2017}  
straightforwardly yields 
\[
\mathbb{P} 
\left[
\exists t,
| 
\widehat{P}^{(t)}_{m,d} 
- 
P_{m,d}
| 
{\geq} 
\sqrt{
\frac{ n^{(t)}_{m} + 1 }
{2}
\ln \frac{ \sqrt{ n^{(t)}_{m} + 1} } {\delta}
}
\frac{1}{n^{(t)}_{m}}
\right]
\leq 
2 \delta.   
\]
Note that $P_{m,d} \in [0,1]$ and $\widehat{P}^{(t)}_{m,d}  \geq 0$.  
Thus the above inequality can be made into 
\[
\mathbb{P} 
\left[
\exists t,
| 
\widehat{P}^{(t)}_{m,d} 
- 
P_{m,d}
| 
{\geq} 
\sqrt{
\frac{ n^{(t)}_{m} + 1 }
{2}
\ln \frac{ \sqrt{ n^{(t)}_{m} + 1} } {\delta}
}
\frac{1}{n^{(t)}_{m}} \wedge 1
\right]
\leq 
2 \delta.   
\]
This is equivalent to 
\[
\mathbb{P} 
\left[
\exists t,
| 
\widehat{P}^{(t)}_{m,d} 
- 
P_{m,d}
| 
{\geq} 
\lambda^{(t)}_{m}
\right]
\leq 
2 \delta.   
\]
The union bounds implies that 
\[
\mathbb{P} 
\left[
\exists t, m, d,
| 
\widehat{P}^{(t)}_{m,d} 
- 
P_{m,d}
| 
{\geq} 
\lambda^{(t)}_{m}
\right]
\leq 
\sum_{ m \in [M], d\in [d_{\max}] } 
\mathbb{P} 
\left[
\exists t,
| 
\widehat{P}^{(t)}_{m,d} 
- 
P_{m,d}
| 
{\geq} 
\lambda^{(t)}_{m}
\right]
\leq 
2 M d_{\max} \delta.   
\]
By a similar argument, one can prove the 
confidence band with respect to the estimator $\widehat{\mu}^{(t)}_{m}$.  
This proof is then complete.

\subsection{Proof of Theorem \ref{thm:InsIndLower}} 

This proof relies on the construction of a special instance of our model.  
The special instance is constructed as follows.   
The move cost of each arm is fixed to zero, formally 
\[
c_{k,m} = 0, \forall k, m.  
\] 
All arms have the same weights, i.e., 
\[
\alpha_1=\alpha_1=\cdots=\alpha_K.  
\]
The reward of each follows a Normal distribution:
\[
R_m \sim \mathcal{N}(\mu_m, \sigma).  
\] 
In each round, each arm generates $K$ units of resource, 
and the number of resource is deterministic, i.e., 
\[
d_{max} = K, P_{m,d} = 1, \forall m,d.  
\]
Suppose that the reward mean satisfies: 
\[
\mu_1 > \mu_2 > \ldots > \mu_M.  
\]
Then the optimal action profile is assigning all plays to arm $m=1$, formally 
\[
a^\ast_k = 1, \forall k.  
\]

Now we prove the regret lower bound based on the above constructed 
special instance of our problem.  
In each round the decision maker needs to assign $K$ plays to arms, 
and multiple arms can be assigned to the same arm.  
For each assigned play $a^{(t)}_k$, if it hits the optimal arm, 
i.e., $a^{(t)}_k = a^\ast_k$, it does not incur regret.  
If it hit a suboptimal arm, say $a^{(t)}_k = m$, where $m \neq 1$, 
it  incurs an expected regret of $\mu_1 - \mu_m$.  
Thus in each round, the action profile $\boldsymbol{a}^{(t)}$ incurs an expected regret of 
\[
\sum_{k \in [K]}  
\alpha_1 (\mu_1 - \mu_{a^{(t)}_k}).  
\]  
Note that this is not MP-MAB, as multiple plays is allowed be assigned to the same arm 
and generates multiple independent rewards with the same distribution associated  with 
this arm.   
We construct a relaxed variant of this problem to make it easier for learning.  
Four model, though the decision maker received $K$ rewards, 
but these reward can be used to update the decision only when this round ends.  
Namely, the decision maker can only update the decision when a round ends.  
To make the learning easier, we allow the decision maker to update the decision 
within each round, when new rewards is received.  
Specifically, we treat each round as $K$ sub-rounds.  
Each sub-round receives one reward and the decision maker can 
update the decision in each sub-round.  
This relaxation improves the data utilization and making the learning easier.  
To unify, we treat each sub-round as a real round.   
This leads to that the decision maker needs to assign one play in each round, 
but in total the decision maker needs to play $KT$ rounds.    
In each round of play, the decision maker observes all historical 
rewards before this round.  
Then this becomes a tradition $M$-armed bandit problem 
with $KT$ rounds of play.  
Applying Theorem 15.2 of \cite{lattimore2020bandit}, 
this problem has a regret lower bound of 
\[
\frac{1}{27} \alpha_1 \sigma \sqrt{M KT}.  
\] 
This proof is then complete.

\subsection{Proof of Theorem \ref{thm:InsDenLower}}

We construct the same instance of MSB-PRS as 
that in the proof of Theorem \ref{thm:InsIndLower}, 
except that we adding more elements to its reward mean.   
The reward mean satisfies: 
\[
\mu_{2} = \mu_1 - \Delta/\alpha_1, \ldots, \mu_{M} = \mu_1 - \Delta/\alpha_1.  
\]
Note that the optimal assign is $a^\ast_k = 1$, namely 
\[ 
U(\boldsymbol{a}^\ast; \boldsymbol{\mu}, \boldsymbol{P})  
= K \alpha_1 \mu_1. 
\]
And the most favorable sub-optimal action profiles are the ones 
that only one play misses the arm $m=1$, 
and all other plays hit the arm $m=1$.  
This implies that  
\[ 
\max_{
\boldsymbol{a}: 
U(\boldsymbol{a}; \boldsymbol{\mu}, \boldsymbol{P}) 
\neq 
U(\boldsymbol{a}^\ast; \boldsymbol{\mu}, \boldsymbol{P})
} 
U(\boldsymbol{a}; \boldsymbol{\mu}, \boldsymbol{P}) 
= 
K \alpha_1 \mu_1 
-
\Delta.  
\]
Thus we have 
\[
U(\boldsymbol{a}^\ast; \boldsymbol{\mu}, \boldsymbol{P})  
-
\max_{
\boldsymbol{a}: 
U(\boldsymbol{a}; \boldsymbol{\mu}, \boldsymbol{P}) 
\neq 
U(\boldsymbol{a}^\ast; \boldsymbol{\mu}, \boldsymbol{P})
} 
U(\boldsymbol{a}; \boldsymbol{\mu}, \boldsymbol{P}) 
=\Delta.  
\]
    
With a similar argument of 
the proof of Theorem \ref{thm:InsIndLower}, 
we only need to analyze the relaxed variant of the instance.  
Note that the relaxed variant is a tradition $M$-armed bandit problem 
with $KT$ rounds of play.  
Applying Theorem 16.2 of \cite{lattimore2020bandit}, 
the asymptotic regret lower bound 
of this variant is 
\[
2 \alpha_1 \sigma^2 \frac{M}{\Delta} \ln KT.  
\]
This proof is then complete.

\subsection{Proof of Theorem \ref{thm:online:InsIndRegret}}

Following the Proof of  \ref{them:InsdeBound}, 
we divide action profiles into groups, such that 
the action profile in the $i$-th group $\mathcal{G}_i$ 
satisfies 
\[
U(\boldsymbol{a}^\ast; \boldsymbol{\mu}, \boldsymbol{P}) - 2^{i} \Delta 
< 
U(a)
\leq 
U(\boldsymbol{a}^\ast; \boldsymbol{\mu}, \boldsymbol{P}) - 2^{i-1} \Delta.  
\]
It has been proved in  the Proof of  \ref{them:InsdeBound}: 
the total number of rounds 
that one profile from group $\mathcal{G}_i$ is selected is upper bounded by  
\[
M \frac{(4 \alpha_1 (\mu_{\max}+1)  \phi(T,\delta)K)^2}{(2^{i-1} \Delta)^2}.  
\] 

Then with a similar argument as the Proof of  \ref{them:InsdeBound} we have: 
\begin{align*}
\text{Reg}_T 
& 
\leq 
2M(1+d_{\max})  K \mu_{\max} 
+ 
\sum_{i \geq \ell} 
M \frac{(4 \alpha_1 (\mu_{\max}+1)  \phi(T,\delta)K)^2}{(2^{i-1} \Delta)^2} 
\times
2^\ell \Delta
+ 
T 2^\ell \Delta 
\\
& 
\leq 
2M(1+d_{\max})  K \mu_{\max} 
+ 
M(4 \alpha_1 (\mu_{\max}+1)  \phi(T,\delta)K)^2  
\sum_{i \geq \ell} 
\frac{1}{(2^{i-1} \Delta)^2} 
\times
2^\ell \Delta
+ 
T 2^\ell \Delta 
\\
& 
\leq 
2M(1+d_{\max})  K \mu_{\max} 
+ 
M(4 \alpha_1 (\mu_{\max}+1)  \phi(T,\delta)K)^2  
\sum_{i \geq \ell} 
\frac{2}{ 2^{i-1} \Delta }  
+ 
T 2^\ell \Delta 
\\
& 
=  
2M(1+d_{\max})  K \mu_{\max} 
+ 
M(4 \alpha_1 (\mu_{\max}+1)  \phi(T,\delta)K)^2  
\sum_{i \geq \ell} 
\frac{4}{ 2^{i} \Delta }  
+ 
T 2^\ell \Delta 
\\
&
=  
2M(1+d_{\max})  K \mu_{\max} 
+ 
M(4 \alpha_1 (\mu_{\max}+1)  \phi(T,\delta)K)^2  
\frac{4}{ 2^{\ell} \Delta }
\sum_{i \geq \ell} 
\frac{1}{ 2^{i-\ell} }  
+ 
T 2^\ell \Delta
\\
& 
\leq 
2M(1+d_{\max})  K \mu_{\max} 
+ 
M (4 \alpha_1 (\mu_{\max}+1)  \phi(T,\delta)K)^2  
\frac{8}{ 2^{\ell} \Delta } 
+ 
T 2^\ell \Delta 
\\
&
\leq 
2M(1+d_{\max})  K \mu_{\max} 
+ 
8\sqrt{8MT} 
 \alpha_1 (\mu_{\max}+1)  \phi(T,\delta)K,  
\end{align*}
where the last step follows by selecting $\Delta$ such that  
\[
M (4 \alpha_1 (\mu_{\max}+1)  \phi(T,\delta)K)^2  
\frac{8}{ 2^{\ell} \Delta } 
=
T 2^\ell \Delta.  
\]
Then by setting $\delta = 1/ T$ we conclude:
\[
\text{Reg}_T 
\leq 
2M(1+d_{\max})  K \mu_{\max} 
+
36
 \alpha_1 (\mu_{\max}+1) (2\sigma +1)  \sqrt{MKT}  \sqrt{K \ln KT} 
\]
This proof is then complete.

\subsection{Proof of Theorem \ref{them:InsdeBound}}

{\bf Technical preparation:} 
Prove the confidence level of the approximate UCB index.  
We aim to prove  the following holds
\[
\mathbb{P} 
\left[
\forall t, \boldsymbol{a}, 
\text{UCB}^{(t)} (\boldsymbol{a})
 - 
U(\boldsymbol{a}; \boldsymbol{\mu}, \boldsymbol{P}) 
 \leq 
\sum_{ m \in [M]}
4 \alpha_1 (\mu_{\max}+1)  
( 
\lambda^{(t)}_{m}
+ 
2 \epsilon^{(t)}_m 
)
\sum\nolimits_{k \in [K]} 
\boldsymbol{1}_{\{ a_{k} = m \}}, 
\atop
\text{UCB}^{(t)} (\boldsymbol{a}) \geq U(\boldsymbol{a}; \boldsymbol{\mu}, \boldsymbol{P})
\right]
\geq 
1 - 2M (1+d_{\max}) \delta.  
\]

Lemma \ref{lem:online:mu} and union bounds imply: 
\[
\mathbb{P}
\left[
\forall t, m, d, 
| 
\mu_m
-
\widehat{\mu}^{(t)}_{m}  
|
\leq
\epsilon^{(t)}_{m}, 
| 
\widehat{P}^{(t)}_{m,d} 
- 
P_{m,d}
| 
{\leq} 
\lambda^{(t)}_{m}
\right] 
\geq 
1 - 2M (1+d_{\max}) \delta.  
\]
We next derive an upper bound of 
$ 
\text{UCB}^{(t)} (\boldsymbol{a})
 - 
U(\boldsymbol{a}; \boldsymbol{\mu}, \boldsymbol{P}) 
$,  
given 
$
| 
\mu_m
-
\widehat{\mu}^{(t)}_{m}  
|
\leq
\epsilon^{(t)}_{m}
$
and 
$ 
| 
\widehat{P}^{(t)}_{m,d} 
- 
P_{m,d}
| 
{\leq} 
\lambda^{(t)}_{m}.  
$
First we have: 
\begin{align*}
\text{UCB}^{(t)} (\boldsymbol{a})
 - 
U(\boldsymbol{a}; \boldsymbol{\mu}, \boldsymbol{P}) 
& 
=
U(\boldsymbol{a}, \widehat{ \boldsymbol{\mu} }^{(t)} {+} \boldsymbol{\epsilon}^{(t)} ,
\widehat{\boldsymbol{P}}^{(t)} 
{+} \boldsymbol{\lambda}^{(t)})
- 
U(\boldsymbol{a}; \boldsymbol{\mu}, \boldsymbol{P}) 
\\
&
= 
\sum_{m \in [M]} 
( 
U_m (\boldsymbol{a}, \widehat{ \mu}^{(t)}_m + \epsilon^{(t)}_m ,
\widehat{\boldsymbol{P}}^{(t)}_m 
{+} \boldsymbol{\lambda}^{(t)}_m)
- 
U_m(\boldsymbol{a}; \mu_m, \boldsymbol{P}_m) 
)
\\
& 
 = \sum_{m \in [M]} 
\Phi^{(t)}_m, 
\end{align*}
where $\Phi^{(t)}_m$ is defined as $\Phi^{(t)}_m \triangleq 
U_m (\boldsymbol{a}, \widehat{ \mu}^{(t)}_m + \epsilon^{(t)}_m ,
\widehat{\boldsymbol{P}}^{(t)}_m 
{+} \boldsymbol{\lambda}^{(t)}_m)
- 
U_m(\boldsymbol{a}; \mu_m, \boldsymbol{P}_m).  
$  
Note that 
\begin{align*}
& 
U_m (\boldsymbol{a}, \widehat{ \mu}^{(t)}_m + \epsilon^{(t)}_m ,
\widehat{\boldsymbol{P}}^{(t)}_m 
{+} \boldsymbol{\lambda}^{(t)}_m) 
=
(\widehat{ \mu}^{(t)}_m + \epsilon^{(t)}_m)
      \sum\nolimits_{k \in [K]} 
      \boldsymbol{1}_{\{ a_{k} = m \}} \alpha_k 
( \widehat{P}^{(t)}_{m,\ell^{(t)}_{k}} + \lambda^{(t)}_{m})
- 
\sum_{k\in [M]} c_{k,m} \boldsymbol{1}_{ \{a_{k}=m\}}, 
\\
& 
U_m(\boldsymbol{a}; \mu_m, \boldsymbol{P}_m)  
=
\mu_m 
      \sum\nolimits_{k \in [K]} 
      \boldsymbol{1}_{\{ a_{k} = m \}} \alpha_k P_{m,\ell^{(t)}_{k}}
- 
\sum_{k\in [M]} c_{k,m} \boldsymbol{1}_{ \{a_{k}=m\}}.
\end{align*}
Note that $
| 
\mu_m
-
\widehat{\mu}^{(t)}_{m}  
|
\leq
\epsilon^{(t)}_{m}
$ 
implies that 
$\widehat{\mu}^{(t)}_{m} \leq 
\mu_m + \epsilon^{(t)}_{m}$
and 
$\widehat{\mu}^{(t)}_{m} +\epsilon^{(t)}_{m}
\geq 
\mu_m$.  
Furthermore, 
$| 
\widehat{P}^{(t)}_{m,d} 
- 
P_{m,d}
| 
{\leq} 
\lambda^{(t)}_{m}
$
implies that 
$
\widehat{P}^{(t)}_{m,d} 
\leq
P_{m,d}
+
\lambda^{(t)}_{m}
$
and 
$
\widehat{P}^{(t)}_{m,d} 
+
\lambda^{(t)}_{m} 
\geq P_{m,d}
$. 
The condition $\widehat{\mu}^{(t)}_{m} \leq 
\mu_m + \epsilon^{(t)}_{m}$ 
and 
$
\widehat{P}^{(t)}_{m,d} 
\leq
P_{m,d}
+
\lambda^{(t)}_{m}
$
yields the following upper bound of 
$U_m (\boldsymbol{a}, \widehat{ \mu}^{(t)}_m + \epsilon^{(t)}_m ,
\widehat{\boldsymbol{P}}^{(t)}_m 
{+} \boldsymbol{\lambda}^{(t)}_m)$ as follows: 
\begingroup
\allowdisplaybreaks
\begin{align*}
& 
U_m 
(
\boldsymbol{a}, 
\widehat{ \mu}^{(t)}_m + \epsilon^{(t)}_m ,
\widehat{\boldsymbol{P}}^{(t)}_m +  \boldsymbol{\lambda}^{(t)}_m
) 
\\
& 
=
(
\widehat{ \mu}^{(t)}_m + \epsilon^{(t)}_m)
      \sum\nolimits_{k \in [K]} 
      \boldsymbol{1}_{\{ a_{k} = m \}} \alpha_k 
( \widehat{P}^{(t)}_{m,\ell^{(t)}_{k}} + \lambda^{(t)}_{m})
- 
\sum_{k\in [M]} c_{k,m} \boldsymbol{1}_{ \{a_{k}=m\}} 
\\
& 
\leq 
(
\mu_m + 2 \epsilon^{(t)}_m
)
      \sum\nolimits_{k \in [K]} 
      \boldsymbol{1}_{\{ a_{k} = m \}} \alpha_k 
( \widehat{P}^{(t)}_{m,\ell^{(t)}_{k}} + \lambda^{(t)}_{m})
- 
\sum_{k\in [M]} c_{k,m} \boldsymbol{1}_{ \{a_{k}=m\}} 
\\
& 
\leq 
(
\mu_m + 2 \epsilon^{(t)}_m
)
\sum\nolimits_{k \in [K]} 
\boldsymbol{1}_{\{ a_{k} = m \}} \alpha_k 
( P_{m,\ell^{(t)}_{k}} + 2 \lambda^{(t)}_{m})
- 
\sum_{k\in [M]} c_{k,m} \boldsymbol{1}_{ \{a_{k}=m\}} 
\\
& 
= 
U_m(\boldsymbol{a}; \mu_m, \boldsymbol{P}_m)  
+ 
\mu_m
\sum\nolimits_{k \in [K]} 
\boldsymbol{1}_{\{ a_{k} = m \}} \alpha_k 
2 \lambda^{(t)}_{m}
+ 
2 \epsilon^{(t)}_m 
\sum\nolimits_{k \in [K]} 
\boldsymbol{1}_{\{ a_{k} = m \}} \alpha_k 
( P_{m,\ell^{(t)}_{k}} + 2 \lambda^{(t)}_{m}) 
\\
& 
\leq 
U_m(\boldsymbol{a}; \mu_m, \boldsymbol{P}_m) 
+ 
2 \alpha_1 \mu_{\max}  \lambda^{(t)}_{m} 
\sum\nolimits_{k \in [K]} 
\boldsymbol{1}_{\{ a_{k} = m \}}
+ 
2 \epsilon^{(t)}_m 
(\alpha_1 + 2  \alpha_1  \lambda^{(t)}_{m})
\sum\nolimits_{k \in [K]} 
\boldsymbol{1}_{\{ a_{k} = m \}}
\\
& 
= U_m(\boldsymbol{a}; \mu_m, \boldsymbol{P}_m) 
+ 
2 \alpha_1 (\mu_{\max}+1)  
( 
\lambda^{(t)}_{m}
+ 
 \epsilon^{(t)}_m
)  
\sum\nolimits_{k \in [K]} 
\boldsymbol{1}_{\{ a_{k} = m \}}
+
4  \alpha_1 
\epsilon^{(t)}_m 
\lambda^{(t)}_{m} 
\sum\nolimits_{k \in [K]} 
\boldsymbol{1}_{\{ a_{k} = m \}}
\\
& 
\leq 
U_m(\boldsymbol{a}; \mu_m, \boldsymbol{P}_m) 
+
 2 \alpha_1 (\mu_{\max}+1)  
( 
\lambda^{(t)}_{m}
+ 
 \epsilon^{(t)}_m
)  
\sum\nolimits_{k \in [K]} 
\boldsymbol{1}_{\{ a_{k} = m \}}
+
4 \alpha_1 (\mu_{\max}+1) 
\epsilon^{(t)}_m 
\lambda^{(t)}_{m} 
\sum\nolimits_{k \in [K]} 
\boldsymbol{1}_{\{ a_{k} = m \}}
\\
& 
\leq  
U_m(\boldsymbol{a}; \mu_m, \boldsymbol{P}_m) 
+
 4 \alpha_1 (\mu_{\max}+1)  
( 
\lambda^{(t)}_{m}
+ 
 \epsilon^{(t)}_m
+ 
\epsilon^{(t)}_m 
\lambda^{(t)}_{m}
)
\sum\nolimits_{k \in [K]} 
\boldsymbol{1}_{\{ a_{k} = m \}} 
\\
& 
\leq 
 U_m(\boldsymbol{a}; \mu_m, \boldsymbol{P}_m) 
+
 4 \alpha_1 (\mu_{\max}+1)  
( 
\lambda^{(t)}_{m}
+ 
2 \epsilon^{(t)}_m 
)
\sum\nolimits_{k \in [K]} 
\boldsymbol{1}_{\{ a_{k} = m \}},    
\end{align*}
\endgroup
where the last step follows $\lambda^{(t)}_{m} \leq 1$.  
Then it follows that 
\[
\Phi^{(t)}_m 
= 
U_m 
(
\boldsymbol{a}, 
\widehat{ \mu}^{(t)}_m + \epsilon^{(t)}_m ,
\widehat{\boldsymbol{P}}^{(t)}_m +  \boldsymbol{\lambda}^{(t)}_m
) 
- 
U_m(\boldsymbol{a}; \mu_m, \boldsymbol{P}_m)
\leq 
4 \alpha_1 (\mu_{\max}+1)  
( 
\lambda^{(t)}_{m}
+ 
2 \epsilon^{(t)}_m
)
\sum\nolimits_{k \in [K]} 
\boldsymbol{1}_{\{ a_{k} = m \}}
.  
\] 
The condition $\widehat{\mu}^{(t)}_{m} +\epsilon^{(t)}_{m}
\geq 
\mu_m$ 
and 
$
\widehat{P}^{(t)}_{m,d} 
+
\lambda^{(t)}_{m} 
\geq P_{m,d}
$ 
yields that 
\[
U_m 
(
\boldsymbol{a}, 
\widehat{ \mu}^{(t)}_m + \epsilon^{(t)}_m ,
\widehat{\boldsymbol{P}}^{(t)}_m +  \boldsymbol{\lambda}^{(t)}_m
) 
\geq 
(
\boldsymbol{a}, 
\mu_m,
\widehat{\boldsymbol{P}}^{(t)}_m +  \boldsymbol{\lambda}^{(t)}_m
) 
\geq 
U_m(\boldsymbol{a}; \mu_m, \boldsymbol{P}_m).  
\]
Where each step is a consequence of the piece-wise monotone 
property of the utility function.  
Summing them up with respect to $m$, step I concludes.

We first give a definition, which is useful in our proof.  
\begin{definition}
An action profile $\boldsymbol{a}$ is $\Psi$-optimal, if it satisfies 
\[
U(\boldsymbol{a}; \boldsymbol{\mu}, \boldsymbol{P}) > 
U(\boldsymbol{a}^\ast; \boldsymbol{\mu}, \boldsymbol{P}) - \Psi. 
\]
where $\Psi \in \mathbb{R}_+$.  
\end{definition}

{\bf Step I: } 
Prove sufficient conditions on $n^{(t)}_m$, 
such that the selected action profile is $\Psi$-optimal. 

Note that 
\[
U(\boldsymbol{a}^\ast; \boldsymbol{\mu}, \boldsymbol{P})
\leq
\text{UCB}^{(t)} (\boldsymbol{a}^\ast) 
\leq 
\text{UCB}^{(t)} (\boldsymbol{a}^{(t)}) 
\leq 
U(\boldsymbol{a}^{(t)}; \boldsymbol{\mu}, \boldsymbol{P}) 
+
\gamma^{(t)} (\boldsymbol{a}^{(t)}) 
\]
This implies that 
\[
U(\boldsymbol{a}^{(t)}; \boldsymbol{\mu}, \boldsymbol{P}) 
\geq 
U(\boldsymbol{a}^\ast; \boldsymbol{\mu}, \boldsymbol{P})
-
\gamma^{(t)} (\boldsymbol{a}^{(t)}).  
\]
Thus in each round $t$, the pulled action profile 
is $\gamma^{(t)} (\boldsymbol{a}^{(t)})$-optimal.  
The $\lambda^{(t)}_m $ can be bounded as
\[
\lambda^{(t)}_m \leq 
\sqrt{
\frac{ n^{(t)}_{m} + 1 }
{2}
\ln \frac{ \sqrt{ n^{(t)}_{m} + 1} } {\delta}
}
\frac{1}{n^{(t)}_{m}}
\leq 
 \sqrt{
\frac{1}{n^{(t)}_{m}}
\ln \frac{ \sqrt{T} } {\delta}
}
\leq 
 \sqrt{ 
\ln \frac{ \sqrt{T} } {\delta}
}
\sqrt{
\frac{1}{n^{(t)}_{m}}
}.  
\]
The $\epsilon^{(t)}_m$  can be bounded as 
\[
\epsilon^{(t)}_m 
=
\sqrt{
2
\sigma^2
(\widetilde{n}^{(t)}_m+1)
\ln \frac{ \sqrt{\widetilde{n}^{(t)}_m +1} }{ \delta }
}
\frac{1}{\widetilde{n}^{(t)}_m}
\leq 
2 \sigma  
\sqrt{ 
\frac{1}{\widetilde{n}^{(t)}_m}
\ln \frac{ \sqrt{KT} }{ \delta }
}
\leq 
2 \sigma 
 \sqrt{
\frac{1}{n^{(t)}_{m}}
\ln \frac{ \sqrt{KT} } {\delta}
}
\leq 
2 \sigma 
 \sqrt{ 
\ln \frac{ \sqrt{KT} } {\delta}
}
 \sqrt{
\frac{1}{n^{(t)}_{m}}
}.
\]
Then it follows that 
\begin{align*}
\gamma^{(t)}(\boldsymbol{a})
& 
=  
\sum_{ m \in [M]}
4 \alpha_1 (\mu_{\max}+1)  
( 
\lambda^{(t)}_{m}
+ 
2 \epsilon^{(t)}_m 
)
\sum\nolimits_{k \in [K]} 
\boldsymbol{1}_{\{ a_{k} = m \}}
\\
&
\leq 
\sum_{ m \in [M]}
4 \alpha_1 (\mu_{\max}+1)  
\sum\nolimits_{k \in [K]} 
\boldsymbol{1}_{\{ a_{k} = m \}} 
\left(
 \sqrt{ 
\ln \frac{ \sqrt{T} } {\delta}
}
\sqrt{
\frac{1}{n^{(t)}_{m}}
}
+
2 \sigma 
 \sqrt{ 
\ln \frac{ \sqrt{KT} } {\delta}
}
 \sqrt{
\frac{1}{n^{(t)}_{m}}
}
\right)
\\
& 
\leq
\sum_{ m \in [M]}
4 \alpha_1 (\mu_{\max}+1)  
\sum\nolimits_{k \in [K]} 
\boldsymbol{1}_{\{ a_{k} = m \}} 
\left(
 \sqrt{ 
\ln \frac{ \sqrt{T} } {\delta}
}
+
2 \sigma 
 \sqrt{ 
\ln \frac{ \sqrt{KT} } {\delta}
} 
\right)
\sqrt{
\frac{1}{n^{(t)}_{m}}
}
\\
& 
<
\sum_{ m \in [M]}
4 \alpha_1 (\mu_{\max}+1)  
\sum\nolimits_{k \in [K]} 
\boldsymbol{1}_{\{ a_{k} = m \}}  
(2 \sigma +1)
 \sqrt{ 
\ln \frac{ \sqrt{KT} } {\delta}
} 
\sqrt{
\frac{1}{n^{(t)}_{m}}
}
\\
& 
= 
\sum_{ m \in [M]}
4 \alpha_1 (\mu_{\max}+1)  
\sum\nolimits_{k \in [K]} 
\boldsymbol{1}_{\{ a_{k} = m \}} 
\phi(T,\delta)
\sqrt{
\frac{1}{n^{(t)}_{m}}
}
\\
& 
=
4 \alpha_1 (\mu_{\max}+1)  \phi(T,\delta)
\sum_{ m \in [M]}
\sum\nolimits_{k \in [K]} 
\boldsymbol{1}_{\{ a_{k} = m \}} 
\sqrt{
\frac{1}{n^{(t)}_{m}}
}, 
\end{align*}
where $\phi(T,\delta)$ is defined as 
\[
\phi(T,\delta) 
\triangleq 
(2 \sigma +1)
 \sqrt{ 
\ln \frac{ \sqrt{KT} } {\delta}
}.
\]
One sufficient condition to guarantee $\gamma^{(t)}(\boldsymbol{a}) \leq \Psi$ is  
\[
n^{(t)}_{m} 
\geq 
\frac{(4 \alpha_1 (\mu_{\max}+1)  \phi(T,\delta)K)^2}{\Psi^2}, 
\forall m \in [M].
\]
This can be shown by 
\begin{align*}
\gamma^{(t)}(\boldsymbol{a})  
& 
< 
4 \alpha_1 (\mu_{\max}+1)  \phi(T,\delta)
\sum_{ m \in [M]}
\sum\nolimits_{k \in [K]} 
\boldsymbol{1}_{\{ a_{k} = m \}} 
\sqrt{
\frac{1}{n^{(t)}_{m}}
}
\\
& 
\leq 
4 \alpha_1 (\mu_{\max}+1)  \phi(T,\delta)
\sum_{ m \in [M]}
\sum\nolimits_{k \in [K]} 
\boldsymbol{1}_{\{ a_{k} = m \}}  
\frac{\Psi}{4 \alpha_1 (\mu_{\max}+1)  \phi(T,\delta)K}
\\
& 
= 
4 \alpha_1 (\mu_{\max}+1)  \phi(T,\delta)K
\frac{\Psi}{4 \alpha_1 (\mu_{\max}+1)  \phi(T,\delta)K}
\\
& 
= 
\Psi.  
\end{align*}

{\bf Step 2: } 
We prove that when confidence bands of parameters hold, 
suboptimal plays make progress in identifying better action profiles.  
Define $\Delta$ as the utility gap between the optimal action profile and 
the least favored sub-optimal action profile, i.e., 
\[
\Delta 
\triangleq 
U(\boldsymbol{a}^\ast; \boldsymbol{\mu}, \boldsymbol{P}) 
-
\max_{
\boldsymbol{a}: 
U(\boldsymbol{a}; \boldsymbol{\mu}, \boldsymbol{P}) 
\neq 
U(\boldsymbol{a}^\ast; \boldsymbol{\mu}, \boldsymbol{P})
} 
U(\boldsymbol{a}; \boldsymbol{\mu}, \boldsymbol{P}).
\]
We divide action profiles into groups, such that 
the action profile in the $i$-th group $\mathcal{G}_i$ 
satisfies 
\[
U(\boldsymbol{a}^\ast; \boldsymbol{\mu}, \boldsymbol{P}) - 2^{i} \Delta 
< 
U(a)
\leq 
U(\boldsymbol{a}^\ast; \boldsymbol{\mu}, \boldsymbol{P}) - 2^{i-1} \Delta.  
\]
Note that the the action profiles in the $0$-th group are $\Delta$-optimal. 
In other words, all action profiles in this group are optimal action profiles.  
We therefore only needs to focus on the groups with index $i\geq 1$.  
In the following analysis, we assume $i \geq 1$.  
Suppose in round $t$, an action profile in $i$-group is selected, 
i.e., 
$
\boldsymbol{a}^{t} \in \arg\max_{\boldsymbol{a} \in \mathcal{A}} \text{UCB}^{(t)} (\boldsymbol{a})
$
and $\boldsymbol{a}^{t} \in \mathcal{G}_i$.  
Then it follows that there exists $m$, 
such that 
\begin{align}
n^{(t)}_{m} 
\leq 
\frac{(4 \alpha_1 (\mu_{\max}+1)  \phi(T,\delta)K)^2}{(2^{i-1} \Delta)^2}.  
\label{Appendix:proof:igroupVio}
\end{align}
Let $\mathcal{M}_i 
\triangleq 
\{
m 
|
n^{(t)}_{m} 
\text{ satisfies Eq. (\ref{Appendix:proof:igroupVio})}
\}
$ 
denote a set of all arms that satisfies 
Eq. (\ref{Appendix:proof:igroupVio}).  
We next prove that at least one arm from the set 
$\mathcal{M}_i $ is pulled, 
i.e., $\boldsymbol{a}^{t}$ makes progress in shrinking these 
inaccurate estimations and thus making progress for 
identifying better action profiles.  
Suppose that none of arms from $\mathcal{M}_i $ are played.  
We next show this leads to a contradiction.  
We construct an instance of the problem by that 
we set all the parameters of the arms to be 
the ground truth and the estimation error to be zero, 
i.e., 
\[
\widehat{\mu}^{(t)}_{m}  
= \mu_m, 
\widehat{P}^{(t)}_{m,d} 
=
P_{m,d},
\epsilon^{(t)}_{m} = 0, 
\lambda^{(t)}_{m} = 0, \forall m \in \mathcal{M}_i. 
\]
We left the parameters of other arms, 
i.e., $m \in [M] \setminus  \mathcal{M}_i$ 
as the estimated ones.  
Let $\text{UCB}^{(t)}_i (\boldsymbol{a}) $ denote the upper confidence of action 
profiles in this constructed instance.  
The first consequence is that this constructed instance 
locating a $2^{i-1} \Delta$ policy, i.e., 
\[
U(\boldsymbol{a}^\ast_i; \boldsymbol{\mu}, \boldsymbol{P}) 
> 
U(\boldsymbol{a}^\ast; \boldsymbol{\mu}, \boldsymbol{P}) 
- 
2^{i-1} \Delta,
\]
where $\boldsymbol{a}^\ast_i \in \arg\max \text{UCB}^{(t)}_i (\boldsymbol{a})$.  
Note that the $\text{UCB}^{(t)}_i (\boldsymbol{a}) $ is piece-wisely increasing and 
the upper confidence of each parameter is larger than its ground truth, 
which yields $\text{UCB}^{(t)}_i (\boldsymbol{a}) \leq \text{UCB}^{(t)} (\boldsymbol{a})$.  
Then it follows that 
\[
\text{UCB}^{(t)} (\boldsymbol{a}^{(t)})
\geq 
\text{UCB}^{(t)} (\boldsymbol{a}^\ast_i)
\geq 
\text{UCB}^{(t)}_i (\boldsymbol{a}^\ast_i) 
\geq 
U(\boldsymbol{a}^\ast_i; \boldsymbol{\mu}, \boldsymbol{P})
>
U(\boldsymbol{a}^\ast; \boldsymbol{\mu}, \boldsymbol{P}) 
- 
2^{i-1} \Delta.  
\]
This contradicts that $\boldsymbol{a}^{(t)}$ belongs to the $i$-th group.  

{\bf Step III:} Regret due to pulling suboptimal arms, 
but the confidence bands of parameters hold.  
In total there are $M$ arms.  
By the argument of Step II, the total number of rounds 
that one profile from group $\mathcal{G}_i$ is selected is upper bounded by  
\[
M \frac{(4 \alpha_1 (\mu_{\max}+1)  \phi(T,\delta)K)^2}{(2^{i-1} \Delta)^2}
\]
rounds.  
Each such play incurs a regret of at most $2^{i} \Delta$.  
Thus the total regret is upper bounded by 
\begin{align*}
\sum_{i=1}^\infty 
M \frac{(4 \alpha_1 (\mu_{\max}+1)  \phi(T,\delta)K)^2}{(2^{i-1} \Delta)^2} 
2^{i} \Delta 
& 
= 
\sum_{i=1}^\infty 
M \frac{(4 \alpha_1 (\mu_{\max}+1)  \phi(T,\delta)K)^2}{(2^{i-1} \Delta)^2} 
2^{i} \Delta 
\\
&
=
M (4 \alpha_1 (\mu_{\max}+1)  \phi(T,\delta)K)^2  
\sum_{i=1}^\infty 
\frac{1}{2^{i-2} \Delta }
\\
&
\leq 
4M (4 \alpha_1 (\mu_{\max}+1)  \phi(T,\delta)K)^2   
\frac{1}{ \Delta }.  
\end{align*}

{\bf Step IV:} The regret due to the failure of confidence bands. 
The regret is upper bounded by 
\[
2M(1+d_{\max}) \delta K \mu_{\max} T.  
\]

{\bf Step V:} The total regret.  
By selecting $\delta= 1/T$, we bound the total regret as: 
\begin{align*}
\text{Reg}_T 
& 
\leq
4M (4 \alpha_1 (\mu_{\max}+1)  \phi(T,1/T)K)^2   
\frac{1}{ \Delta } + 
2M(1+d_{\max})  K \mu_{\max} 
\\
& 
\leq 
64M K^2 \alpha^2_1 (\mu_{\max}+1)^2  (\phi(T,1/T)K)^2   
\frac{1}{ \Delta } + 
2M(1+d_{\max}) K \mu_{\max}  
\\
& 
\leq 
64 M K^2 \alpha^2_1 
\left(
(2 \sigma +1)
 \sqrt{ 
\ln  T \sqrt{KT}  
}
\right)^2
\frac{1}{ \Delta } 
+
2M(1+d_{\max})  K \mu_{\max}
\\
& 
= 
64 M K^2 \alpha^2_1 
(2 \sigma +1)^2
\ln  T \sqrt{KT}  
\frac{1}{ \Delta } 
+
2M(1+d_{\max})  K \mu_{\max}
\\
&
\leq 
96 M K^2 \alpha^2_1 
(2 \sigma +1)^2
\frac{1}{ \Delta } 
\ln  KT 
+
2M(1+d_{\max})  K \mu_{\max}.  
\end{align*}
This proof is then complete.

\end{document}